\documentclass[10pt,twocolumn,letterpaper]{article}

\usepackage[pagenumbers]{cvpr}

\usepackage{arydshln}
\usepackage{placeins}
\usepackage[percent]{overpic}

\definecolor{cvprblue}{rgb}{0.21,0.49,0.74}
\usepackage[pagebackref,breaklinks,colorlinks,allcolors=cvprblue]{hyperref}

\def\paperID{00000}
\def\confName{CVPR}
\def\confYear{2026}

\title{SketchSense: Learning to Interpret Imperfect Sketch Guidance for Image Inpainting}

\author{
Zian Yang\\
Fudan University\\
{\tt\small 23307140016@m.fudan.edu.cn}
}

\begin{document}
\maketitle
\begin{abstract}
Sketch-guided image inpainting provides intuitive structural control, yet real
sketches often mix reliable global intent with locally crowded, displaced,
incomplete, or deliberately unconventional strokes. Existing approaches
typically either retain the input sketch as a fixed condition throughout
denoising or refine it into a clean structure before RGB synthesis. The former
assumes uniformly reliable strokes and can propagate local errors throughout
generation; the latter must resolve ambiguous structure before emerging
appearance and semantic context become available. We propose \textbf{SketchSense},
a framework that interprets imperfect sketch guidance by synchronously denoising
interacting RGB and structure streams. Bidirectional Attention Fusion couples
appearance generation with structural recovery, producing a refined structure
that exposes the model's evolving sketch interpretation. A phrase-level
objective aligns the semantic grounding of the two streams. Sketch-Aware
Spatial Regulation further adapts sketch use to local generation states by
modulating attention and the fusion process, while an optional signed prior
injects preserve-versus-correct intent into feature representations and
attention behavior. Experiments on natural and structurally complex imagery
show substantial gains over existing methods in both restoration quality and
structural fidelity.
\end{abstract}

\section{Introduction}
\label{sec:intro}

\begin{figure}[!t]
    \centering
    \includegraphics[page=1,width=\columnwidth]{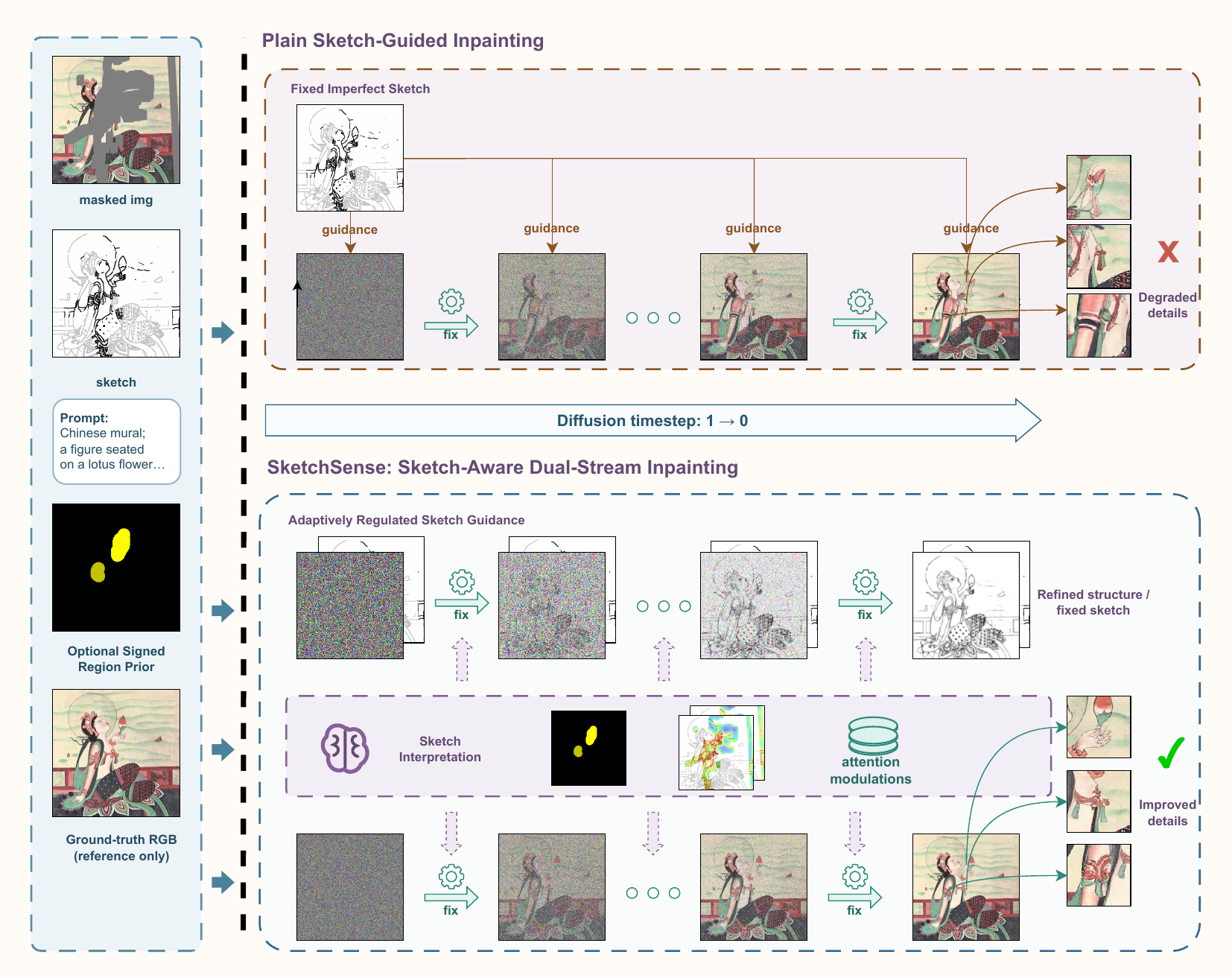}
    \caption{Motivation and overview of SketchSense. Fixed-sketch conditioning
    can propagate locally degraded strokes. SketchSense synchronously denoises
    RGB and structure, regulates sketch use through learned attention
    modulations and an optional signed prior, and exposes its interpretation as
    refined structure.}
    \label{fig:teaser}
\end{figure}

Image inpainting synthesizes coherent content in masked regions while
preserving compatibility with the observed image. Sketches complement prompts
and masks by directly specifying spatial layout, object boundaries, and local
artistic intent. Their sparse, editable form lets users mark critical contours
without providing a complete structural map. Prior work has demonstrated their
value for interactive, shape-guided, and structure-aware
inpainting~\cite{Liu2024SketchRefiner, Sharma2024SketchGuided, Xie2023SmartBrush,
Liu2024StructureMatters}, particularly when plausible restoration depends on
fine contours, ornamental patterns, or complex structural organization.

The key challenge is that a user sketch is neither a uniformly accurate target
nor merely a weak hint. It often captures the intended global layout and useful
boundaries, while local strokes may be crowded, displaced, incomplete, or
distorted~\cite{Liu2024SketchRefiner, Sharma2024SketchGuided}. Conversely, an
unconventional shape may reflect deliberate artistic intent. Treating every
stroke as equally reliable can transfer drawing errors to the restored image,
whereas globally weakening the condition can discard structures the user wants
to preserve. Sketch reliability is therefore spatially heterogeneous, and
geometry
alone may not distinguish an error from deliberate design. The model must
instead relate the sketch to evolving appearance, structure, and semantics,
then adapt its use across locations. This need is especially acute in murals,
artworks, and illustrations, where dense contours and organized patterns
amplify local interpretation errors~\cite{Li2024Mural, Cao2023ZITSpp}.

As illustrated in Fig.~\ref{fig:teaser}, a fixed sketch can propagate errors
from locally degraded strokes, whereas SketchSense interprets and regulates the
same guidance during synchronous denoising.

Many conditional generation pipelines use a sketch, edge map, or line drawing
as a fixed external condition throughout
denoising~\cite{Zhang2023ControlNet, Tan2025OminiControl}. Spatial weighting and
attention control can vary where or how strongly the condition is
read~\cite{Kim2023DenseDiffusion, Cohen2025ConditionalBalance}, but the
underlying sketch remains fixed. These controls rescale its influence without
exposing an evolving interpretation, so local errors may persist throughout
denoising.

Another strategy first recovers a clean structure and then synthesizes RGB.
Although this structure-first formulation exposes an intermediate result and
is common in sketch refinement and structure-prior
inpainting~\cite{Liu2024SketchRefiner, Li2024Mural, Cao2023ZITSpp}, it must resolve
structural ambiguity before appearance cues such as color, material, and
texture emerge. Errors in that estimate are then inherited by the RGB generator
as fixed conditions. This is restrictive in structurally complex imagery, where the
meaning of a stroke and its visual realization are closely coupled. These
limitations motivate recovering appearance and structure together.

We present \textbf{SketchSense}, a sketch-guided inpainting framework built
around synchronous RGB--structure understanding. Two interacting branches
jointly denoise RGB and structure latents: the RGB branch reconstructs the
image, while the structure branch turns the imperfect sketch into refined
structure at the same denoising step. \textit{Bidirectional Attention
Fusion} exchanges time-aligned information, allowing appearance and text
context to inform structure recovery while progressively refined structure
guides RGB synthesis. The refined-structure trajectory exposes the model's
interpretation of the input sketch and supports inspection or iterative
revision.

During training, modality differences can cause the same object to be localized
differently by text-conditioned attention in the RGB and structure branches.
We therefore introduce \textit{Phrase-Level Cross-Branch Semantic Consistency},
an auxiliary objective that aligns their phrase-conditioned spatial attention
while retaining modality-specific representations.

Locally heterogeneous sketch evidence is not equally useful at every location
or transformer block. Global scales and fixed region weights cannot capture
this variation, especially because blocks respond differently to
conditioning~\cite{Cohen2025ConditionalBalance}. We therefore introduce
\textit{Intrinsic Sketch-Aware Spatial Regulation}, which uses evolving
dual-branch states and local geometry to modulate attention according to sketch
reliability and gate the fusion process across space and blocks.

Sketch geometry alone may not reveal whether an unconventional stroke should be
preserved or corrected. \textit{Explicit Signed Sketch-Use Modeling} addresses
this ambiguity with an optional signed spatial prior that communicates local
intent. The prior modulates token representations, sketch-source attention, and
retrieved sketch features.


In summary, our contributions are threefold:
\begin{itemize}
    \item We propose a synchronous RGB--structure framework for
    interpreting imperfect sketch guidance. Bidirectional Attention Fusion and
    phrase-level semantic consistency couple the two branches, which produce
    restored RGB and an observable refined structure.
    \item We introduce Intrinsic Sketch-Aware Spatial Regulation, which learns
    to adapt sketch attention and the fusion process across locations and
    transformer blocks.
    \item We introduce Explicit Signed Sketch-Use Modeling to express local
    preservation-or-correction intent when geometry alone is ambiguous, using
    an optional prior to regulate sketch representation and use.
\end{itemize}

\section{Related Work}
\label{sec:related_work}

\subsection{Sketch- and Structure-Guided Image Inpainting}

Sketch- and structure-guided inpainting augments masks and text with visual
conditions that specify missing geometry. SmartBrush combines text with shape
guidance for object inpainting, while Structure Matters introduces auxiliary
structure to reduce semantic discrepancy in diffusion
inpainting~\cite{Xie2023SmartBrush, Liu2024StructureMatters}. Together, these
methods establish structural conditions as effective interfaces for
controllable restoration. When the input structure is imperfect, SketchRefiner
refines a user sketch before inpainting, and
structure-prior pipelines recover line or edge representations before color and
detail synthesis~\cite{Liu2024SketchRefiner, Li2024Mural, Cao2023ZITSpp}.
These approaches generally treat structure either as a condition to follow or
as a completed intermediate result. The former does not distinguish reliable
strokes from ambiguous local evidence, whereas the latter resolves ambiguity
before appearance cues emerge. Neither formulation directly addresses a single
sketch whose reliability and intended use vary across regions. SketchSense
instead interprets structure jointly with RGB recovery and retains the refined
structure as an observable result.

\subsection{Joint Multimodal Generation and Semantic Coupling}

Joint diffusion models couple RGB with dense modalities through interacting
branches~\cite{Zhang2023JointNet, Kwon2025JointDiT}, while diffusion-transformer
control methods process image and condition tokens jointly for flexible visual
conditioning~\cite{Tan2025OminiControl}. Related text--region grounding methods
associate textual concepts with spatial support: attention refocusing improves
text--layout correspondence through cross- and self-attention
control~\cite{Phung2024AttentionRefocusing}, and MTADiffusion combines edge
prediction with mask--text alignment for object
inpainting~\cite{Huang2025MTADiffusion}. These works establish effective
multimodal interaction and spatial semantic alignment. Generic multimodal
coupling, however, does not ensure that two modality branches ground the same
concept in the same spatial region, while existing grounding mechanisms
typically align text and layout within one generation path. SketchSense focuses
on this cross-branch mismatch: concurrently denoised RGB and structure branches
may localize the same object phrase differently. It therefore combines
bidirectional feature exchange with an auxiliary phrase-level objective that
aligns spatial grounding across the two modalities.

\subsection{Spatially Adaptive Conditional Control}

Spatial control methods regulate where and how strongly a condition affects
diffusion generation. S-CFG replaces a global classifier-free guidance scale
with semantic-region-dependent guidance~\cite{Shen2024SCFG}, Conditional
Balance studies layer-dependent sensitivity to
conditioning~\cite{Cohen2025ConditionalBalance}, and attention modulation
provides region-specific control over intermediate attention
maps~\cite{Kim2023DenseDiffusion}. These methods provide useful mechanisms for
spatial adherence, condition strength, and layer selection, generally assuming
that the intended meaning of the condition is already known. Imperfect sketches
add a distinct ambiguity because a local deviation may be an error to correct
or an intentional shape to preserve. The model must infer the reliability of
sketch evidence, while the user may also wish to specify its intended use.
SketchSense addresses both cases through intrinsic state-dependent regulation
and an optional signed spatial prior, without treating sketch reliability as a
single global quantity.

\begin{figure*}[!t]
    \centering
    \includegraphics[page=4,width=\textwidth]{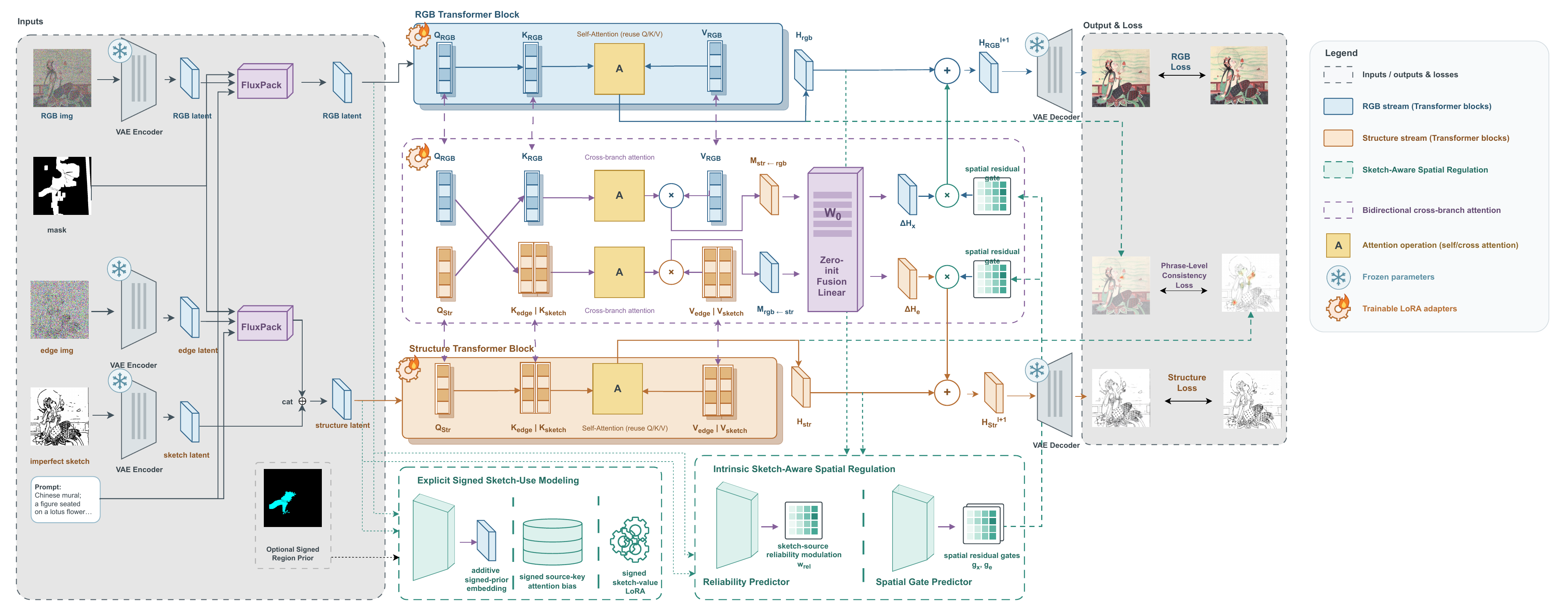}
    \caption{Overview of SketchSense. Synchronous RGB and structure branches
    exchange information through Bidirectional Attention Fusion. Spatial
    regulation adapts sketch attention and fusion injection, and an optional
    signed prior expresses local preservation or correction intent. Training
    uses edge supervision, phrase consistency, and branch losses; inference
    returns restored RGB and refined structure.}
    \label{fig:method_overview}
\end{figure*}

\section{Method}
\label{sec:method}

Figure~\ref{fig:method_overview} summarizes the inference
path and training objectives.
SketchSense synchronously denoises RGB and structure streams, exchanges their
evolving evidence through Bidirectional Attention Fusion, and spatially
regulates sketch use. The RGB stream restores appearance, while the structure
stream interprets imperfect guidance as refined structure.

\subsection{Synchronous RGB--Structure Understanding}
\label{sec:synchronous_understanding}

Let $z_{x,t}$ and $z_{e,t}$ be the noisy RGB and edge latents at the shared
timestep, $x$ and $e$ the visible conditions, $m$ the hole mask, $p$ the
prompt, and $z_s$ the sketch latent. Their initial image-token states are
\begin{equation}
\begin{aligned}
H_x^0&=\mathcal P_x(z_{x,t},x,m,p),\\
H_{\mathrm{str}}^0&=[H_e^0;H_s^0]\\
&=[\mathcal P_e(z_{e,t},e,m,p);\mathcal P_s(z_s)].
\end{aligned}
\label{eq:branch_packing}
\end{equation}
Here $\mathcal P_x$ and $\mathcal P_e$ abstract native FLUX Fill preparation,
projection, and prompt conditioning, $\mathcal P_s$ packs and projects the
sketch, and the semicolon denotes token-axis concatenation.
Sketch tokens remain available as source evidence throughout denoising, while
the edge-target tokens evolve under both structural and appearance context at
every transformer block.
Edge supervision is used during training; at inference, this trajectory records
the model's progressive interpretation of the imperfect sketch.

\subsubsection{Bidirectional Attention Fusion}
\label{sec:bidirectional_fusion}

\begin{figure}[t]
    \centering
    \includegraphics[page=1,width=\columnwidth]{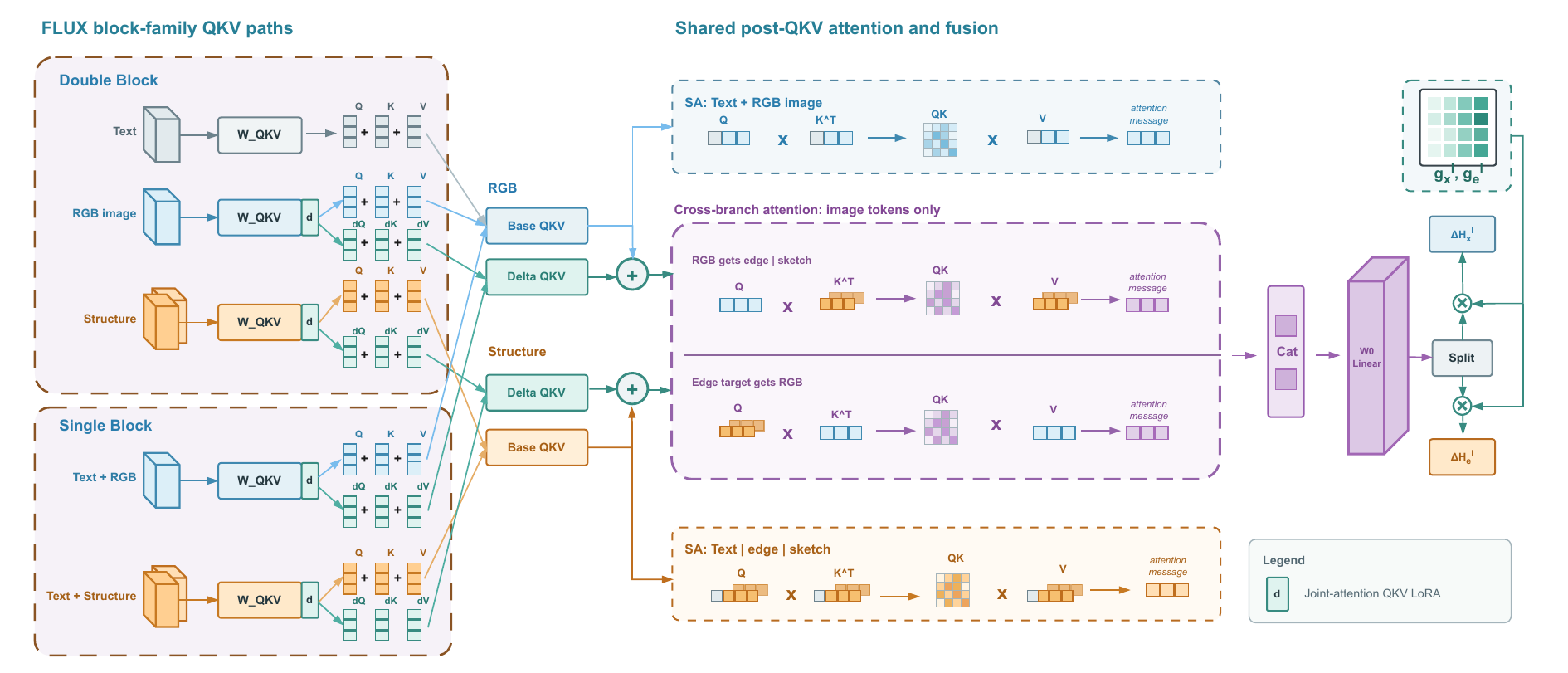}
    \caption{Bidirectional Attention Fusion with adapted QKV reuse. Native QKV
    provides the shared base, and a dedicated joint-attention QKV LoRA produces
    fusion-specific $\Delta QKV$. Directional messages pass through a
    zero-initialized projection, split into $\Delta H_x^l$ and $\Delta H_e^l$,
    and enter their receiving branches through $g_x^l$ and $g_e^l$.}
    \label{fig:bidirectional_fusion}
\end{figure}

As shown in Fig.~\ref{fig:bidirectional_fusion}, each fusion block reuses its
native image-token QKV and adds a dedicated joint-attention QKV
LoRA~\cite{Hu2022LoRA}. For $b\in\{x,\mathrm{str}\}$,
\begin{equation}
(\bar Q_b^l,\bar K_b^l,\bar V_b^l)
=\operatorname{QKV}_b^l(H_b^l)+\mathcal L_b^l(H_b^l),
\label{eq:adapted_qkv_reuse}
\end{equation}
where $H_{\mathrm{str}}^l=[H_e^l;H_s^l]$ and $\mathcal L_b^l$ supplies the
fusion-specific $\Delta QKV$. Using the edge-target query span $\bar Q_e^l$
and the full structure-side key/value span, the block forms
\begin{equation}
\begin{aligned}
M_{x\leftarrow \mathrm{str}}^l
  &=\operatorname{Attn}\!\left(\bar Q_x^l,
    \bar K_{\mathrm{str}}^l,\bar V_{\mathrm{str}}^l\right),\\
M_{e\leftarrow x}^l
  &=\operatorname{Attn}\!\left(\bar Q_e^l,\bar K_x^l,\bar V_x^l\right).
\end{aligned}
\label{eq:bidirectional_attention}
\end{equation}
RGB queries therefore retrieve the current edge hypothesis together with the
original sketch evidence, and edge queries retrieve appearance and semantic
context from the RGB state. Sharing the native projections retains their
pretrained representation, while the low-rank residual specializes the
cross-branch retrieval at each block.

The two directional messages are jointly projected and split into
branch-specific candidate residuals:
\begin{equation}
(\Delta H_x^l,\Delta H_e^l)
=\operatorname{Split}\!\left(W_0^l
 [M_{x\leftarrow \mathrm{str}}^l;
 M_{e\leftarrow x}^l]\right).
\label{eq:fusion_residual}
\end{equation}
The zero-initialized $W_0^l$ preserves the pretrained starting behavior;
Sec.~\ref{sec:intrinsic_reliability} controls the spatial injection of both
candidate residuals. When branch resolutions differ, exchanged features are
spatially aligned. The same construction is used in the double-
and single-stream FLUX block families, with their respective native projections
and dedicated low-rank adapters.

\subsubsection{Phrase-Level Cross-Branch Semantic Consistency}
\label{sec:phrase_coupling}

Shared prompts can still yield different phrase grounding across branches. We
extract local noun phrases $\mathcal P(p)=\operatorname{LLM}(p)$, match each
phrase to its prompt tokens, and aggregate its attention over the matched span
and heads into $A_{b,k}^l$. The phrase-conditioned spatial distributions and their
discrepancy are
\begin{align*}
\pi_{b,k}^l&=\operatorname{Norm}\!\left(\operatorname{Pool}_b(A_{b,k}^l)\right),
\qquad b\in\{x,e\},
\\
d_{lk}&=\operatorname{JS}(\pi_{x,k}^l\Vert\pi_{e,k}^l),
\end{align*}
Here $\operatorname{Pool}_b$ aggregates each branch response onto the common
spatial support used by the loss, and
$\operatorname{Norm}(a)=\operatorname{softmax}(\operatorname{zscore}(a)/T)$.
Let $c_{b,lk}=1-\mathcal{H}(\pi_{b,k}^l)/\log N$ be the normalized-entropy
confidence over the $N$ positions of the common support, and let
$c_{lk}=\max(c_{x,lk},c_{e,lk})$. The soft weight and objective are
\begin{align}
\omega_{lk}&=\sigma\!\left(\frac{\operatorname{sg}(d_{lk})-\mu}{\tau}\right)
\nonumber\\[-2pt]
&\quad\cdot\left[c_0+c_1\operatorname{clip}\!\left(\frac{c_{lk}}{c_{\mathrm{ref}}},0,1\right)\right],
\nonumber\\
\mathcal L_{\mathrm{phrase}}
&=\frac{1}{|\Omega|}\sum_{(l,k)\in\Omega}\omega_{lk}d_{lk}.
\label{eq:phrase_loss}
\end{align}
$\Omega$ contains successfully matched, numerically valid block--phrase pairs.
Parser confidence screening removes unreliable phrases, while $\omega_{lk}$
emphasizes mismatched, spatially confident pairs without changing the JS
gradient through its stop-gradient term.

\subsection{Sketch-Aware Spatial Regulation}
\label{sec:spatial_regulation}

\begin{figure}[t]
    \centering
    \includegraphics[page=1,width=\columnwidth]{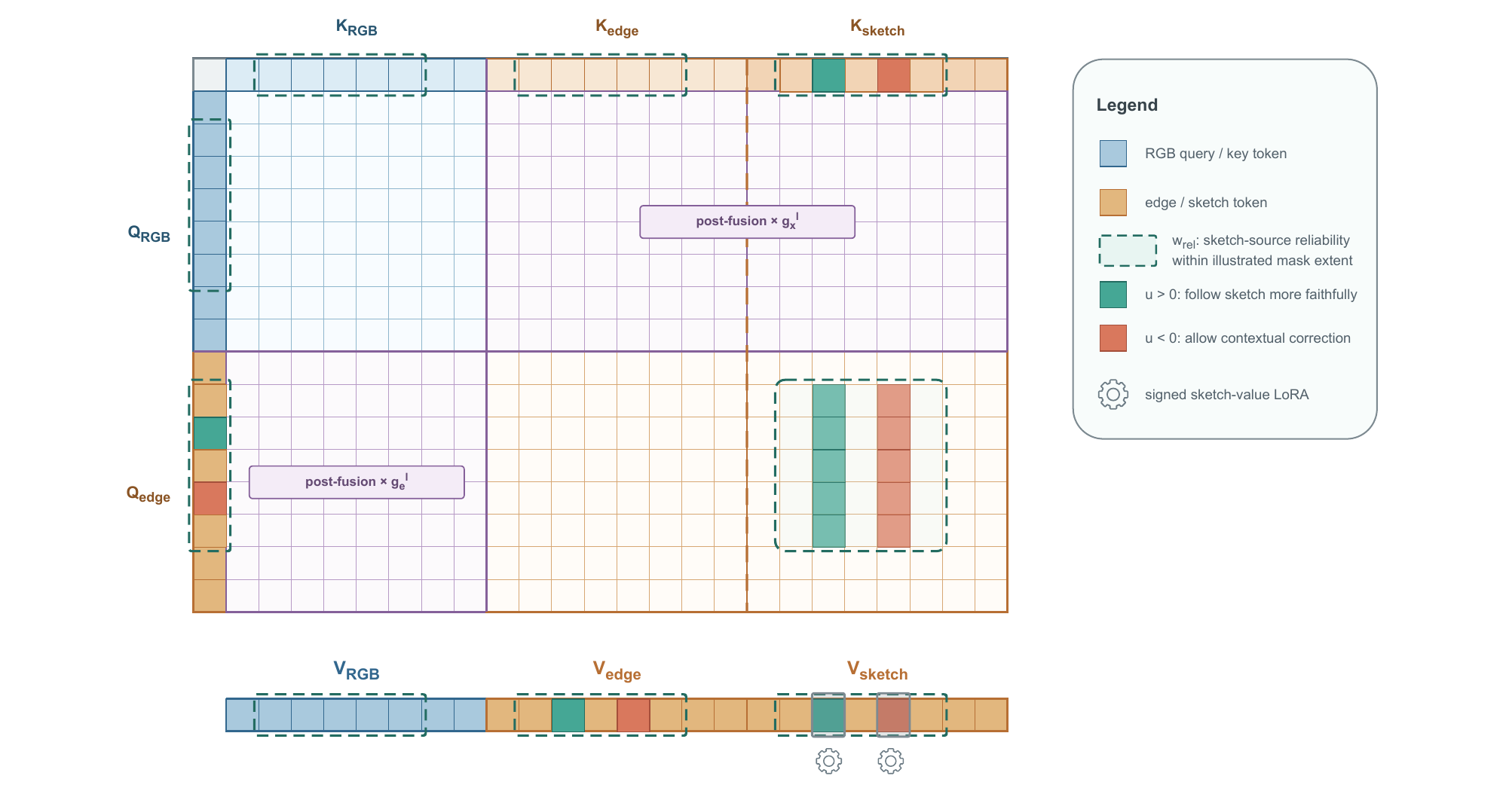}
    \caption{Locations of sketch-aware modulation. The reliability map
    $w_{\mathrm{rel}}$ and signed signal $u$ regulate sketch-source attention
    and value adaptation. Spatial gates $g_x^l$ and $g_e^l$ scale post-fusion
    residuals; colors schematically mark
    mechanism roles.}
    \label{fig:attention_modulations}
\end{figure}

As summarized in Fig.~\ref{fig:attention_modulations}, SketchSense combines
intrinsic state-conditioned regulation with an optional explicit signed prior.
Together, they modulate sketch-source attention and cross-branch residual
injection; their mechanisms are detailed below.

\subsubsection{Intrinsic Sketch-Aware Spatial Regulation}
\label{sec:intrinsic_reliability}

\begin{figure}[t]
    \centering
    \includegraphics[page=1,width=\columnwidth]{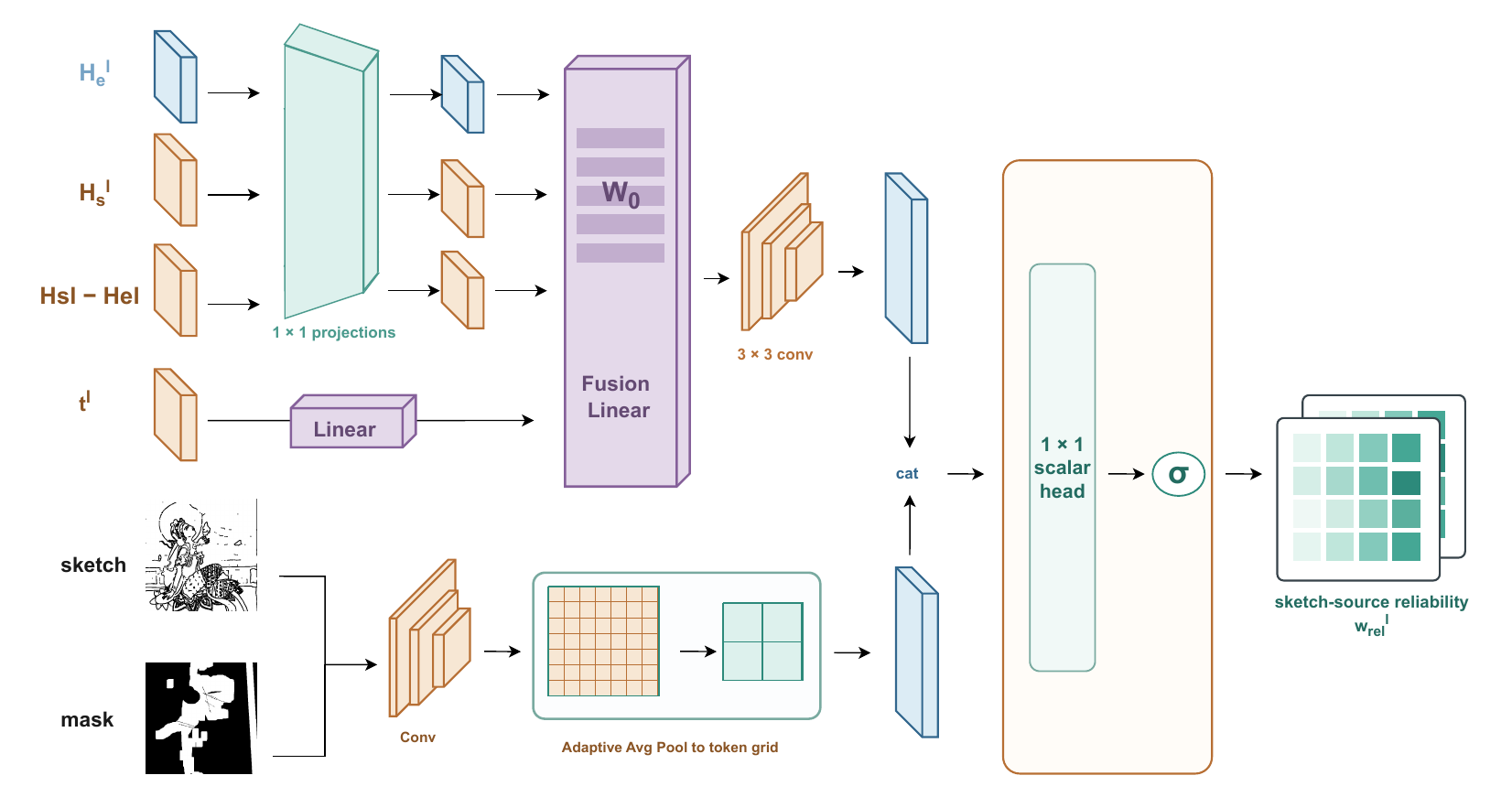}
    \par\smallskip
    \textbf{(a)} Sketch-source reliability predictor.
    \par\medskip
    \includegraphics[page=2,width=\columnwidth]{figs/drawio_sources/exported_figs/figure_05_spatial_regulation_modules.drawio.pdf}
    \par\smallskip
    \textbf{(b)} Spatial residual-gate predictor.
    \caption{Intrinsic Sketch-Aware Spatial Regulation. (a) Edge--sketch
    states and raw geometry produce source reliability
    $w_{\mathrm{rel}}^l$. (b) Branch state, candidate residual, mask tokens, and
    global conditioning produce the residual gate $g_b^l$ for
    $b\in\{x,e\}$.}
    \label{fig:intrinsic_predictors}
\end{figure}

Figure~\ref{fig:intrinsic_predictors} details the reliability and residual-gate
predictors. Let $H_e^l$ and $H_s^l$ be paired edge and sketch states, $S$ the
raw sketch, $M$ the hole mask, and $t^l$ the timestep embedding. Reliability
combines a token-state path with a pixel-space geometry path:
\begin{align}
z_{\mathrm{state}}^l&=\Phi_{\mathrm{state}}^l
\left(H_e^l,H_s^l,H_s^l-H_e^l,t^l\right),
\nonumber\\
z_{\mathrm{geo}}^l&=Z_0^l\!\left(\mathcal P_{h_l,w_l}
\left(\Phi_{\mathrm{geo}}^l([S,M,S\odot M])\right)\right),
\nonumber\\
a^l&=W_{\mathrm{out},0}^l(z_{\mathrm{state}}^l+z_{\mathrm{geo}}^l),
\nonumber\\
w_{\mathrm{rel}}^l&=
\exp\!\left(\log\alpha_{\max}\cdot\tanh(a^l)\right).
\label{eq:intrinsic_reliability}
\end{align}
The adapter $Z_0^l$ and output $W_{\mathrm{out},0}^l$ are zero-initialized, so
$w_{\mathrm{rel}}^l$ starts at one. With edge-side hole-mask tokens $m^l$, source
reliability is confined to restoration queries and supported sketch sources:
\begin{equation}
\bar w_j^l=1+m_j^l(w_{\mathrm{rel},j}^l-1),\qquad
\ell_{ij}^{e\leftarrow s,l}\leftarrow
\ell_{ij}^{e\leftarrow s,l}+m_i^l\log\bar w_j^l.
\label{eq:reliability_attention_bias}
\end{equation}
The token-state path measures agreement between the current edge interpretation
and the sketch. The pixel-space path extracts fine sketch and mask geometry
before adaptive pooling $\mathcal P_{h_l,w_l}$ converts it into token features
for the current transformer block. The bounded exponential converts their
combined evidence into a positive multiplicative reliability: values above one
promote sketch retrieval and values below one attenuate it.

For fusion residual $\Delta H_b^l$, branch state $H_b^l$, mask tokens $m_b^l$,
and global conditioning $v^l$, the block-specific injection is
\begin{align}
g_b^l&=\sigma\!\left(G_{\mathrm{res}}^l
\left(H_b^l,\Delta H_b^l,m_b^l,v^l\right)\right),
\nonumber\\
O_b^l&=O_{b,\mathrm{self}}^l+g_b^l\odot\Delta H_b^l.
\label{eq:spatial_residual_gate}
\end{align}
where $b\in\{x,e\}$ and $O_{b,\mathrm{self}}^l$ is the native update. On the
structure side, $\Delta H_e^l$ has support on the edge-target token span, while
the sketch tokens retain their source role along this residual path. Thus
reliability controls sketch-source selection, whereas $g_b^l$ controls how much
of the candidate cross-branch residual is injected. The gate jointly observes
the receiving state and proposed message, while the mask and global vector
provide restoration support and timestep--prompt conditioning.

\subsubsection{Explicit Signed Sketch-Use Modeling}
\label{sec:explicit_signed_modeling}

\begin{figure}[t]
    \centering
    \includegraphics[page=1,width=\columnwidth]{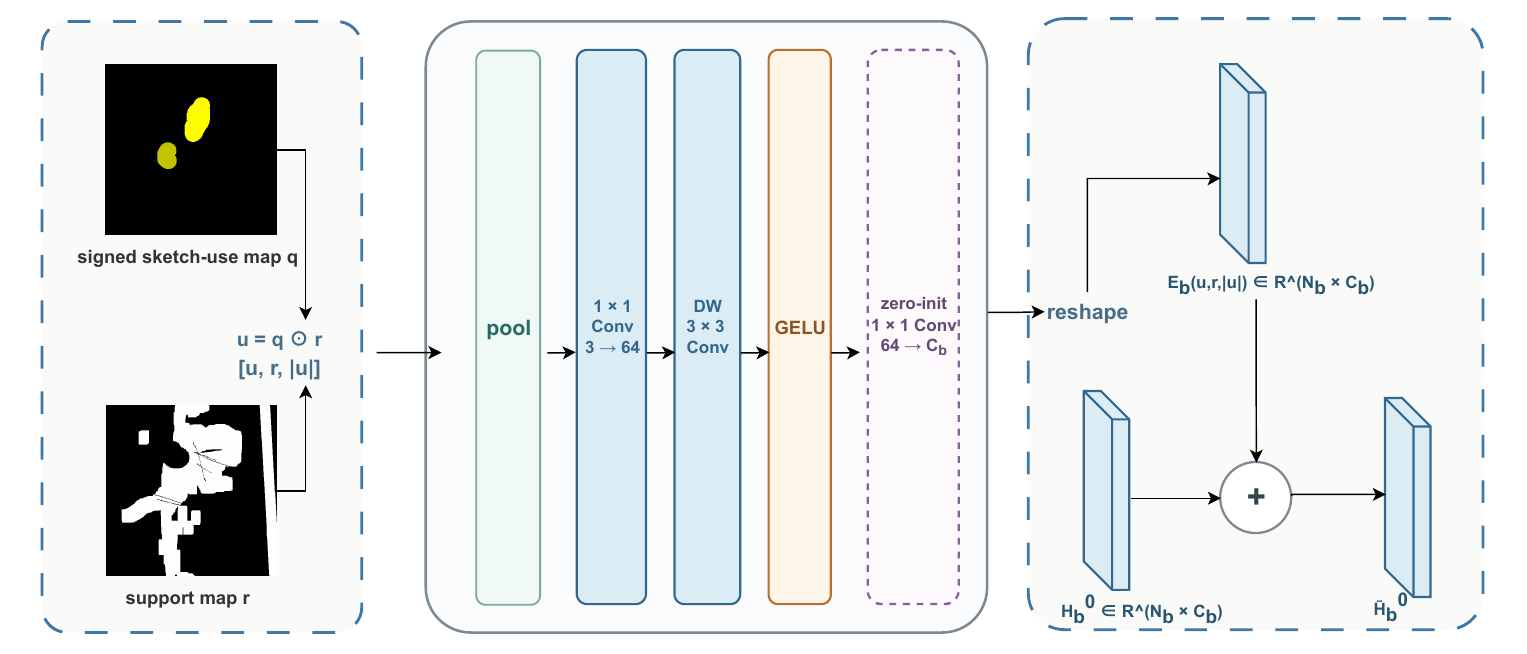}
    \caption{Additive signed-prior embedding. With $u=q\odot r$, a
    zero-initialized encoder maps $[u,r,|u|]$ to the branch-specific $E_b$, which
    forms $\widetilde H_b^0=H_b^0+E_b$. The support map $r$ localizes prior
    activation.}
    \label{fig:signed_prior_embedding}
\end{figure}

Figure~\ref{fig:signed_prior_embedding} illustrates the explicit prior. Let
$q\in[-1,1]^{H\times W}$ express signed sketch-use intent and
$r\in[0,1]^{H\times W}$ its support, with $u=q\odot r$. Positive, negative,
and zero $u$ encourage sketch preservation, permit contextual correction, and
indicate neutral use, respectively. A
zero-initialized spatial encoder adds a branch-specific prior embedding during
initial token preparation:
\begin{equation}
\widetilde H_b^0=H_b^0+E_b(u,r,|u|),\qquad b\in\{x,e,s\}.
\label{eq:prior_embedding}
\end{equation}
The encoder converts $[u,r,|u|]$ into branch-specific token embeddings and
begins with $E_b=0$.
Here $u$ carries supported direction and strength, $r$ records prior support,
and the hole mask $m$ supplies restoration support. Subsequent block-wise
signed modulation is applied through the source-key bias and sketch-value LoRA
defined below.

For restoration branch $b\in\{x,e\}$, query $i$, sketch source $j$, and head
$h$, the prior adjusts source selection by
\begin{equation}
\ell_{b,ij}^{l,h}\leftarrow \ell_{b,ij}^{l,h}
+m_{b,i}^l\,\beta_b^{l,h}u_j,\qquad
\beta_b^{l,h}=\beta_{\max}\tanh(\bar\beta_b^{l,h}).
\label{eq:source_key_bias}
\end{equation}
The signed term raises or lowers supported sketch-key logits with a learned
layer-, head-, and branch-specific sensitivity. It consequently affects both
the RGB path reading structure and the edge path reading sketch evidence. The
bounded coefficient $\beta_b^{l,h}$ keeps this modulation within a fixed scale
while allowing different heads to respond differently to the prior.

The value path applies intent-specific LoRA residuals to the normalized and
modulated sketch hidden state $\widehat H_s^l$ before splitting it into heads:
\begin{equation}
\begin{aligned}
\Delta V_s^l&=\omega_+(u)B_+^lA_+^l\widehat H_s^l
+\omega_-(u)B_-^lA_-^l\widehat H_s^l,\\
\widetilde V_s^{l,h}&=V_s^{l,h}+\operatorname{Split}_h(\Delta V_s^l),
\end{aligned}
\label{eq:value_lora}
\end{equation}
where $\omega_+(u)=\max(u,0)$ and $\omega_-(u)=\max(-u,0)$. The embedding,
source-key bias, and value transformation expose the same signed intent to
token state, sketch selection, and retrieved sketch content, respectively.
Separate positive and negative low-rank factors permit preservation and
correction regions to transform the sketch contribution differently, while
neutral regions retain the base value path.

\subsection{Training Example Construction}
\label{sec:training_construction}

\begin{figure}[t]
    \centering
    \includegraphics[page=2,width=\columnwidth]{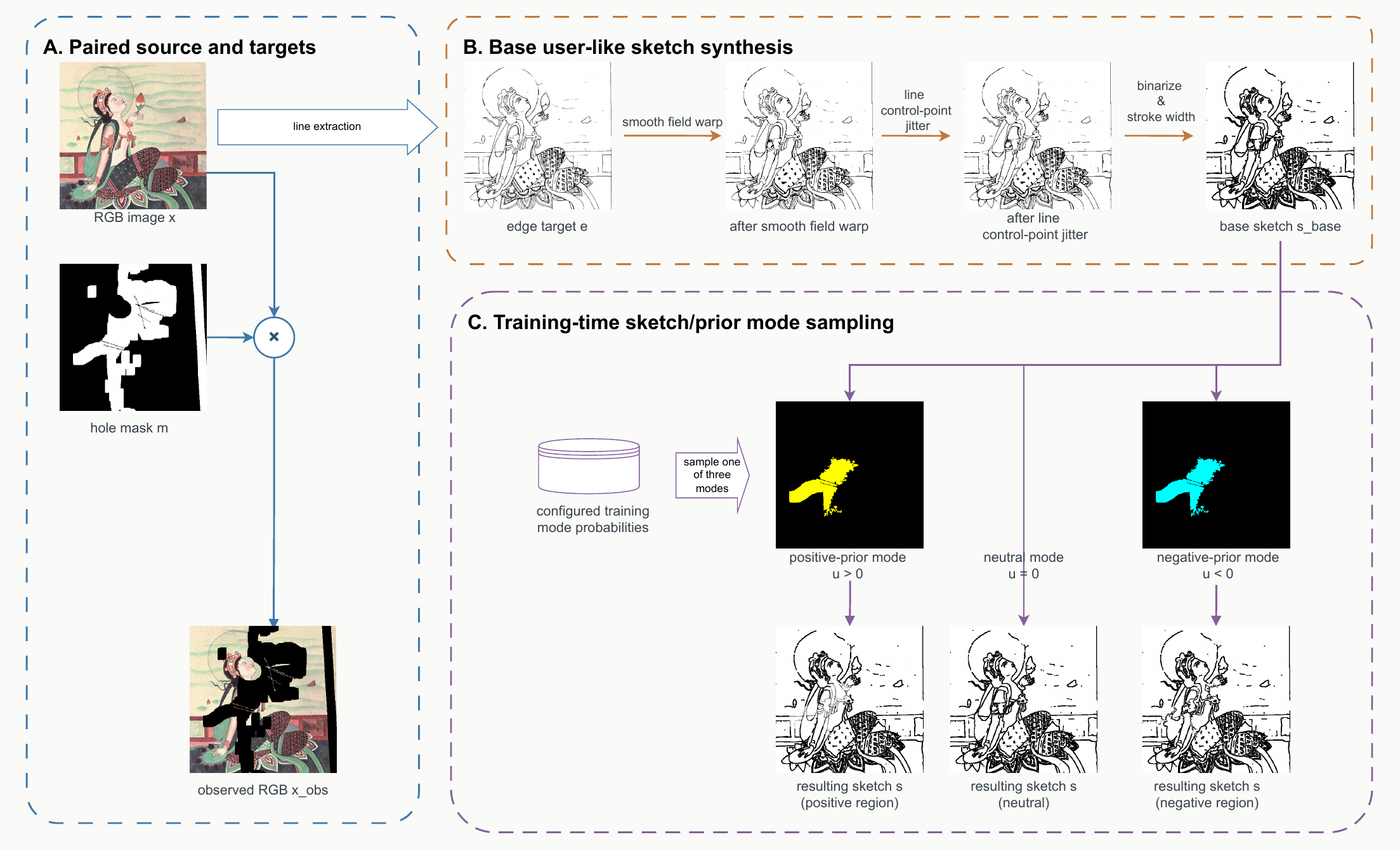}
    \caption{Training example synthesis. Paired RGB and edge targets produce a
    base sketch, followed by one sampled mode: faithful $u>0$, neutral $u=0$,
    or correction-oriented $u<0$. The inpainting mask selects restoration
    pixels, and $r$ selects signed-prior support.}
    \label{fig:training_synthesis}
\end{figure}

We use paired edge annotations when available and otherwise extract edge targets
with MuGE~\cite{Zhou2024MuGE}. From each target, we synthesize a base user-like
sketch primarily through smooth deformation fields, following prior sketch
augmentation practice~\cite{Liu2024SketchRefiner, Sharma2024SketchGuided}. We
then select a local region and sample one treatment: a positive mode replaces it
with more faithful target-derived structure, a negative mode applies stronger warping
with optional stroke removal or an unrelated donor patch, and a neutral mode
keeps the base sketch. These modes yield $u>0$, $u<0$, and $u=0$ training
examples, respectively, while preserving the remaining sketch context, as shown
in Fig.~\ref{fig:training_synthesis}.

\subsection{Training Objective}
\label{sec:training_objective}

For RGB and edge latents $z_b$, $b\in\{x,e\}$, flow
matching~\cite{Lipman2023FlowMatching, Esser2024RectifiedFlowTransformer} uses
noise $\epsilon_b\sim\mathcal{N}(0,I)$ and schedule $\sigma_t$:
\begin{equation}
z_{b,t}=(1-\sigma_t)z_b+\sigma_t\epsilon_b,\qquad
v_b^\star=\epsilon_b-z_b.
\label{eq:flow_matching}
\end{equation}
Both streams share timestep $t$ and predict velocities $\hat v_b$. With binary
restoration mask $m$ and prior support $r$, let $m_{b,i}$ and $r_{b,i}$ be their
pooled values at branch token $i$. The spatial weight and normalized branch
objective are
\begin{equation}
\begin{aligned}
W_{b,i}&=1+\eta_m m_{b,i}+\eta_r r_{b,i},\\
\mathcal L_b&=\mathbb E\!\left[
\frac{\sum_i W_{b,i}\|\hat v_{b,i}-v_{b,i}^\star\|_2^2}
{D_b\sum_i W_{b,i}}
\right],
\end{aligned}
\label{eq:velocity_loss}
\end{equation}
where $D_b$ is the token feature dimension. The RGB term supervises appearance
restoration, the edge term supervises refined structure, and both operate at the
same timestep. Together with Eq.~\eqref{eq:phrase_loss}, the complete objective
is
\begin{equation}
\mathcal{L}=
\frac{\lambda_x\mathcal{L}_x+\lambda_e\mathcal{L}_e}
{\lambda_x+\lambda_e}+
\lambda_{\mathrm{phrase}}\mathcal{L}_{\mathrm{phrase}}.
\label{eq:total_loss}
\end{equation}
Loss weights and sampling ratios are fixed throughout training.

\section{Experiments}
\label{sec:experiments}

\begin{figure*}[t!]
    \centering
    \scriptsize
    \setlength{\tabcolsep}{1.1pt}
    \newcommand{\qualimage}[1]{\includegraphics[width=.087\textwidth,height=.075\textwidth,keepaspectratio]{#1}}
    \newcommand{\qualrowlabel}[1]{\raisebox{.033\textwidth}{#1}}
    \newcommand{\qualmethodlabel}[1]{\raisebox{.010\textwidth}{#1}}
    \begin{tabular}{@{}lccccccccc@{}}
        \toprule
        & \multicolumn{3}{c}{DhMural1714}
        & \multicolumn{3}{c}{ArtBench}
        & \multicolumn{3}{c}{COCO} \\
        \hdashline
        \qualrowlabel{\quad Input: Masked RGB}
        & \qualimage{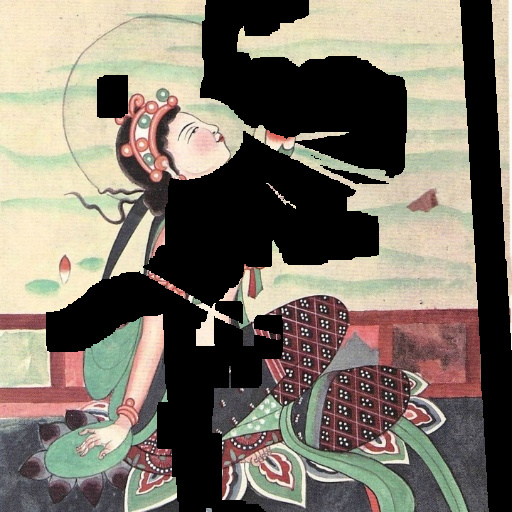}
        & \qualimage{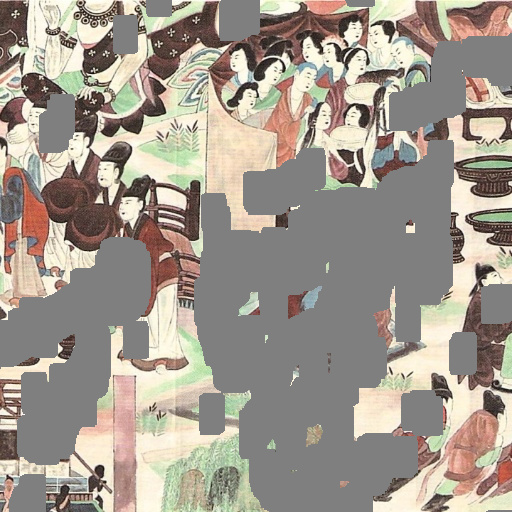}
        & \qualimage{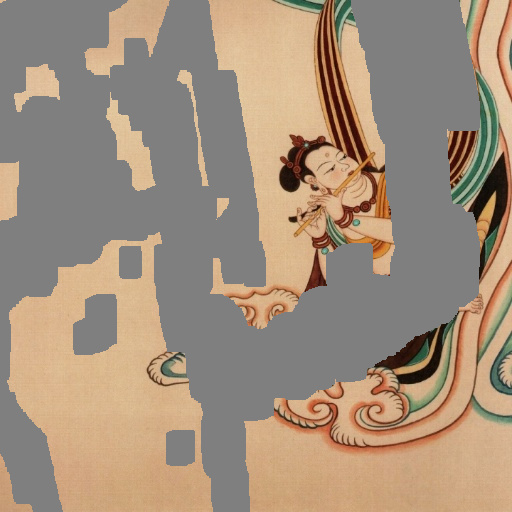}
        & \qualimage{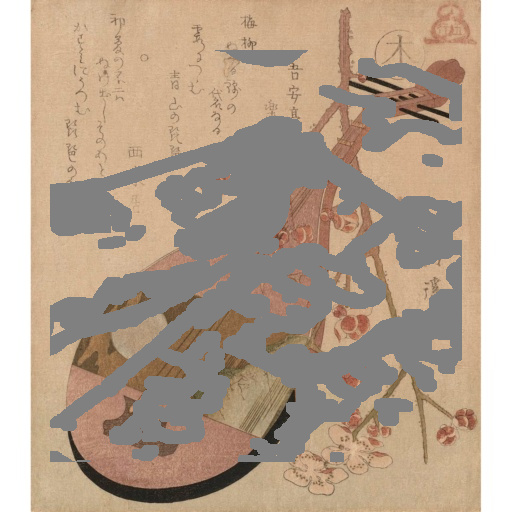}
        & \qualimage{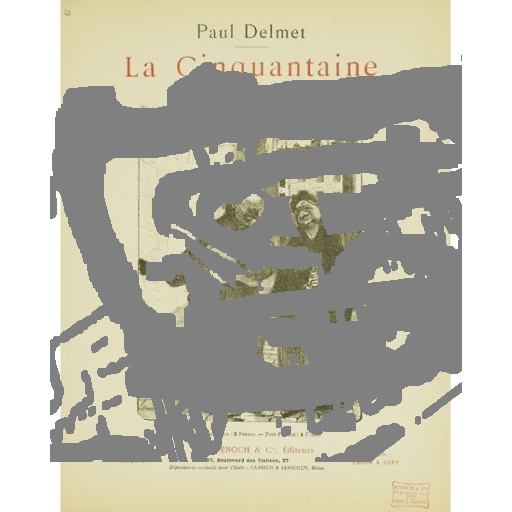}
        & \qualimage{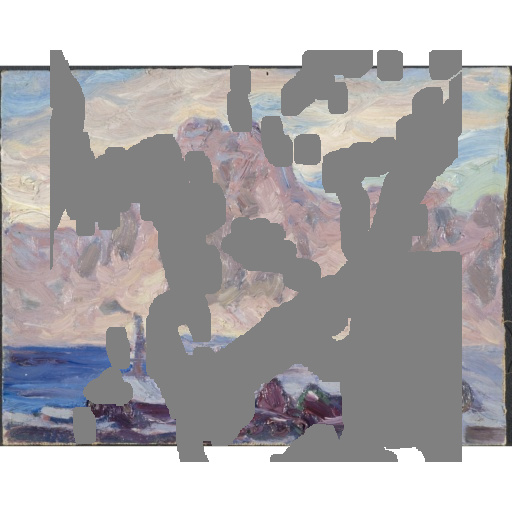}
        & \qualimage{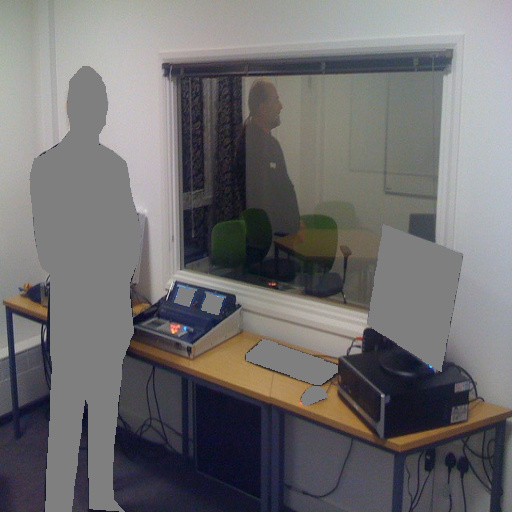}
        & \qualimage{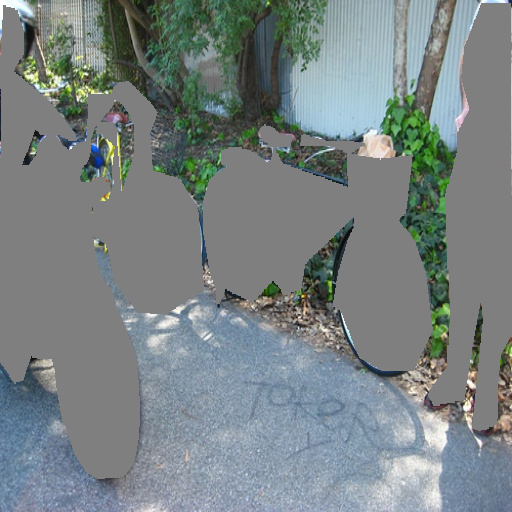}
        & \qualimage{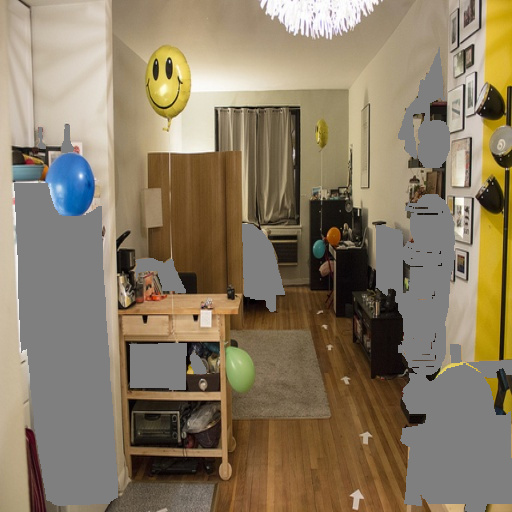} \\
        \qualrowlabel{\quad Input: Sketch}
        & \qualimage{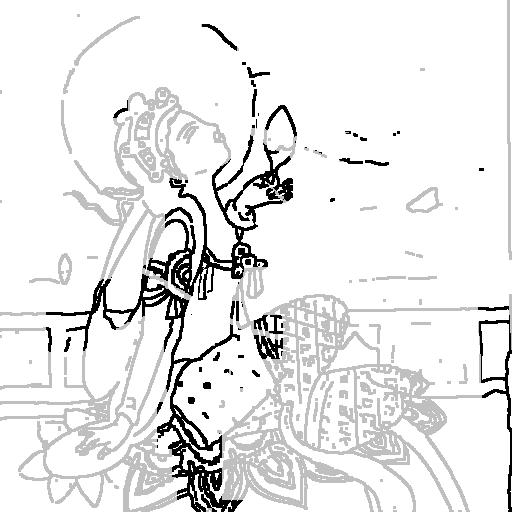}
        & \qualimage{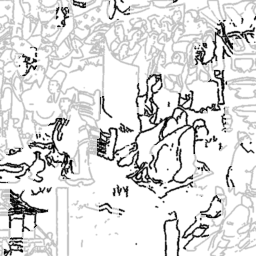}
        & \qualimage{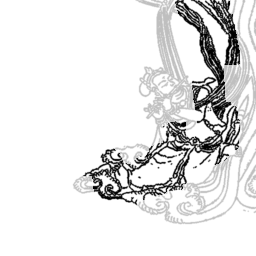}
        & \qualimage{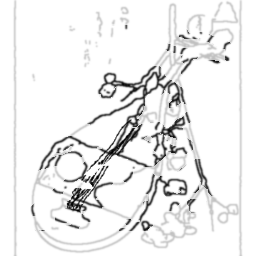}
        & \qualimage{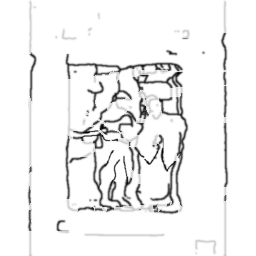}
        & \qualimage{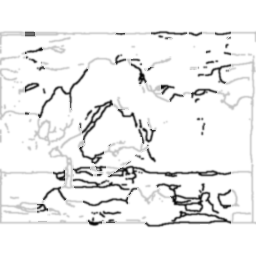}
        & \qualimage{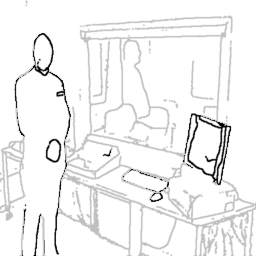}
        & \qualimage{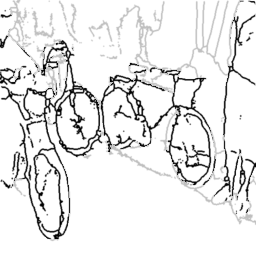}
        & \qualimage{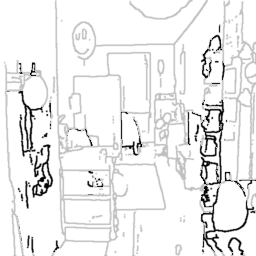} \\
        \hdashline
        \qualmethodlabel{\quad PDDP}
        & \qualimage{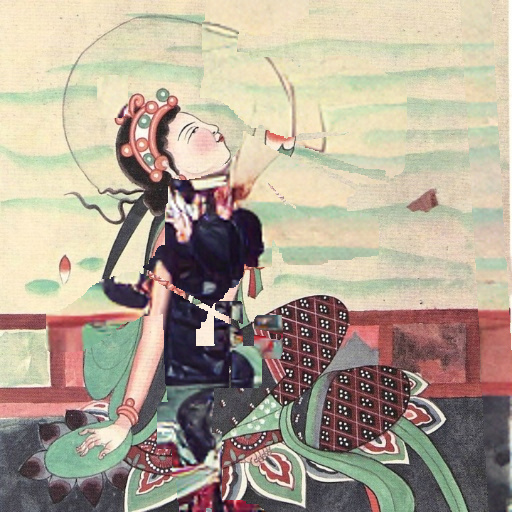}
        & \qualimage{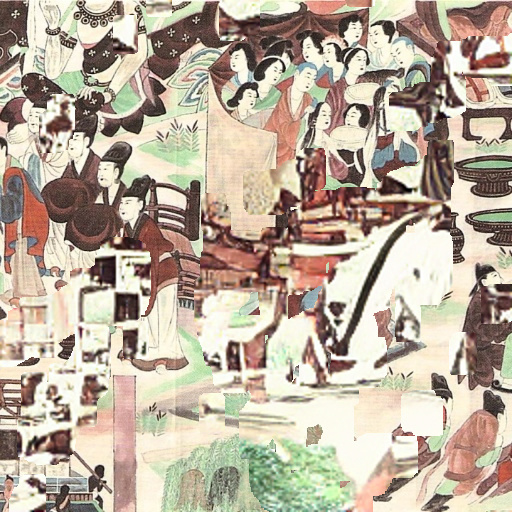}
        & \qualimage{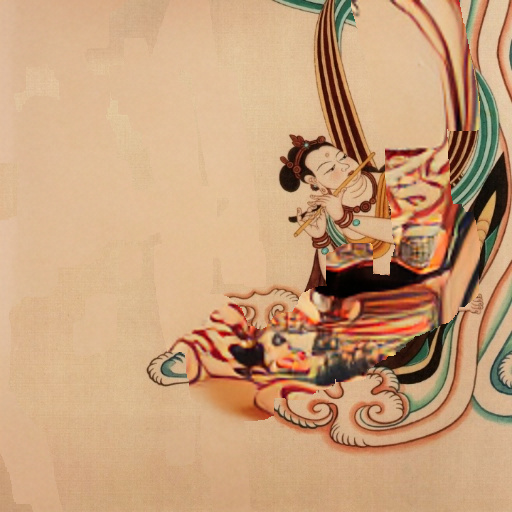}
        & \qualimage{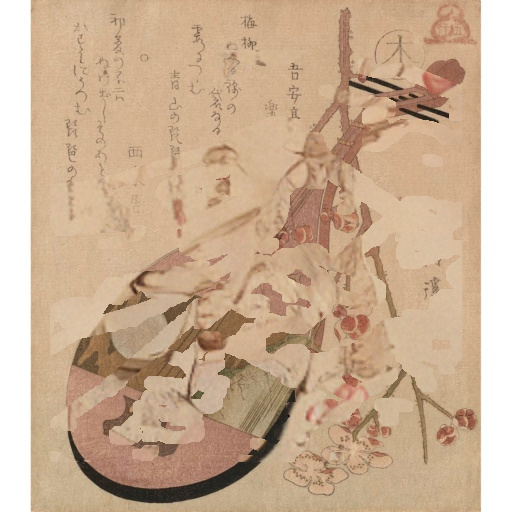}
        & \qualimage{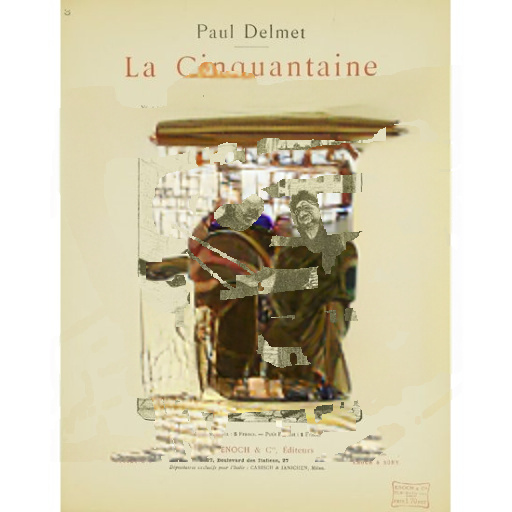}
        & \qualimage{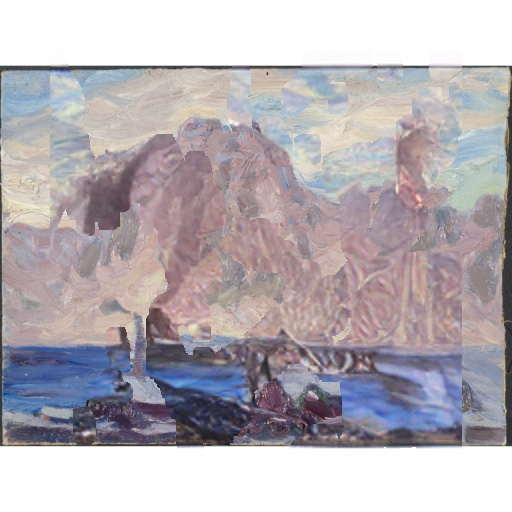}
        & \qualimage{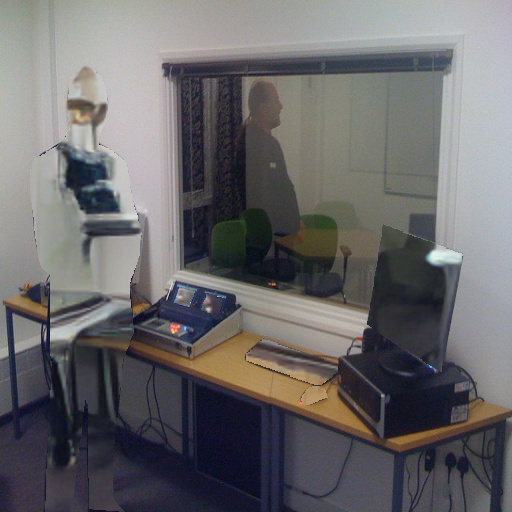}
        & \qualimage{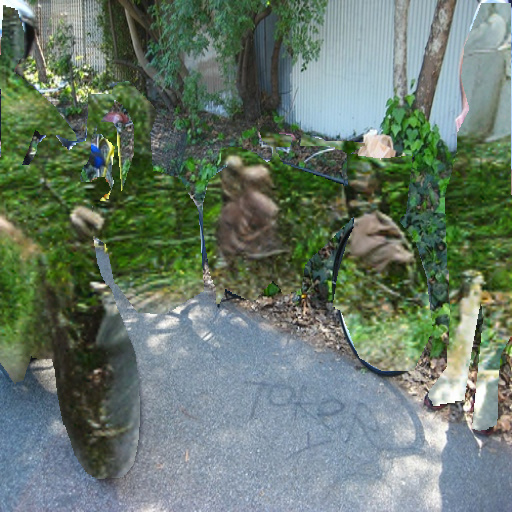}
        & \qualimage{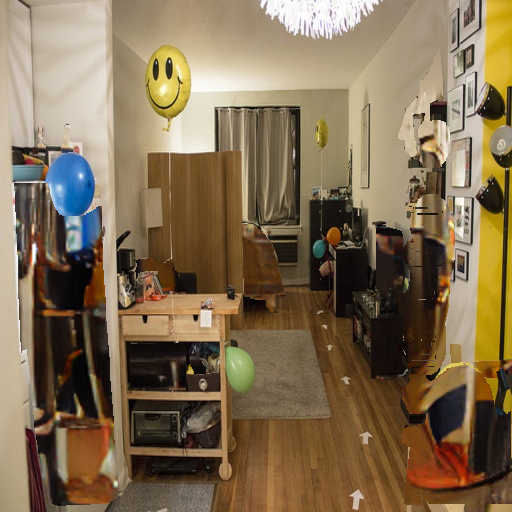} \\
        \qualmethodlabel{\quad MaGIC (adapted)}
        & \qualimage{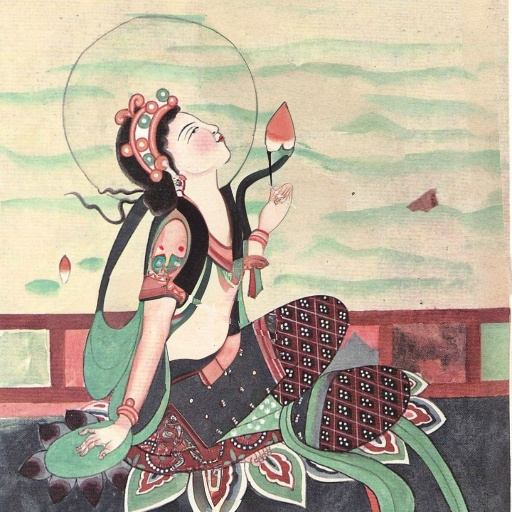}
        & \qualimage{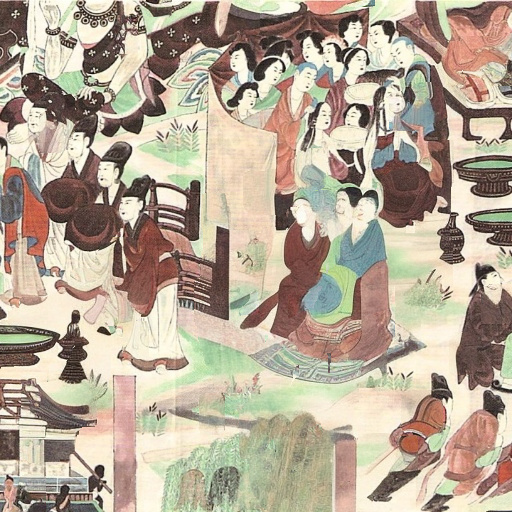}
        & \qualimage{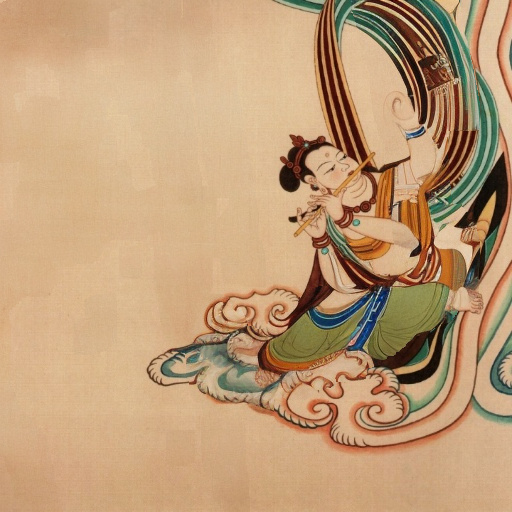}
        & \qualimage{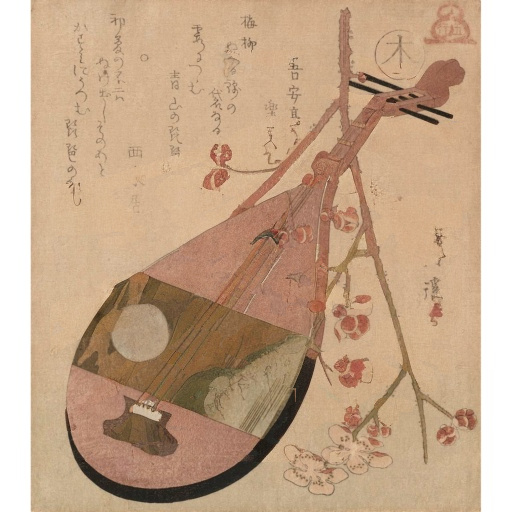}
        & \qualimage{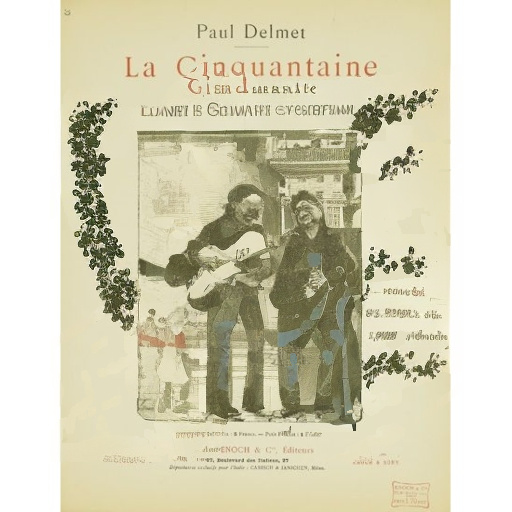}
        & \qualimage{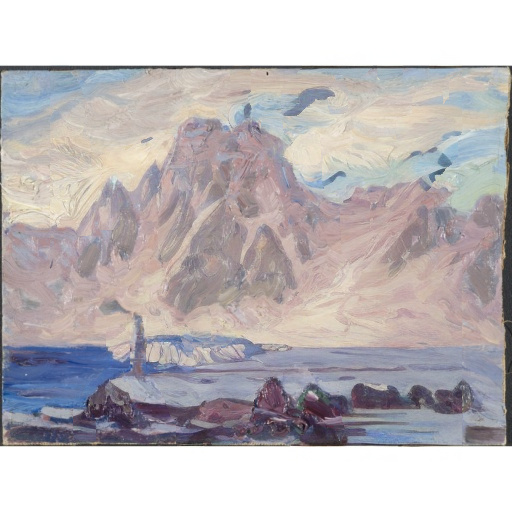}
        & \qualimage{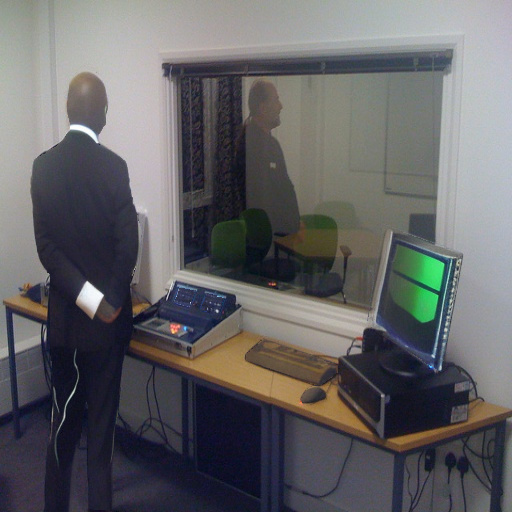}
        & \qualimage{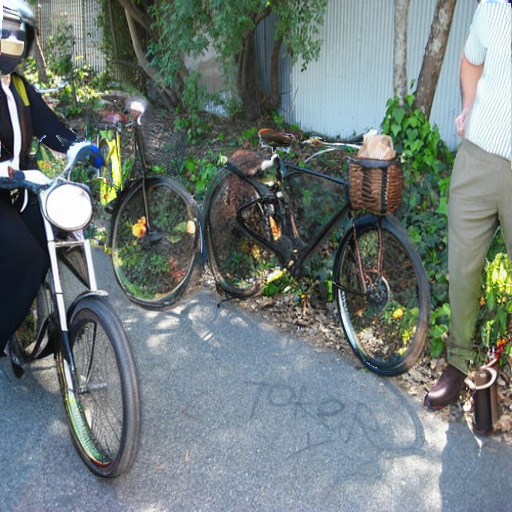}
        & \qualimage{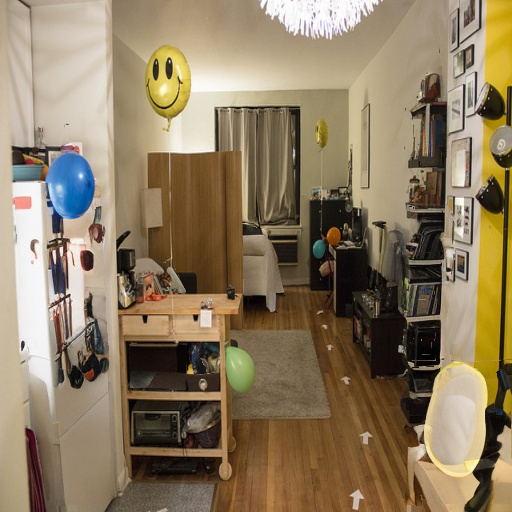} \\
        \qualmethodlabel{\quad SketchRefiner (adapted)}
        & \qualimage{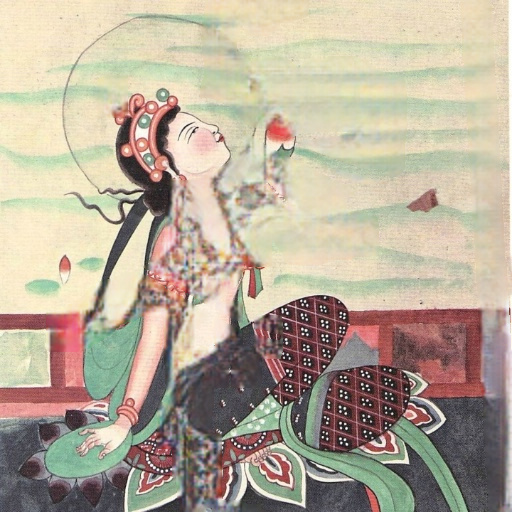}
        & \qualimage{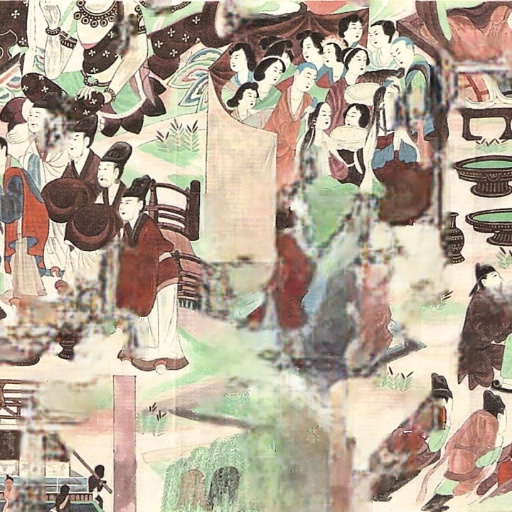}
        & \qualimage{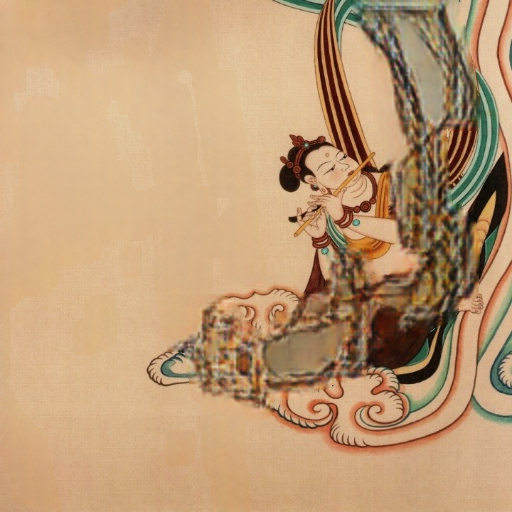}
        & \qualimage{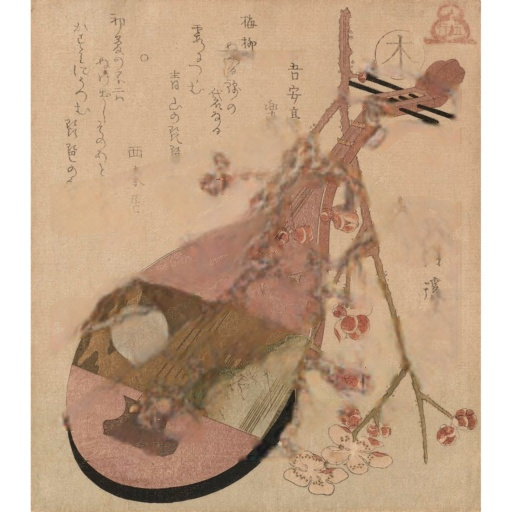}
        & \qualimage{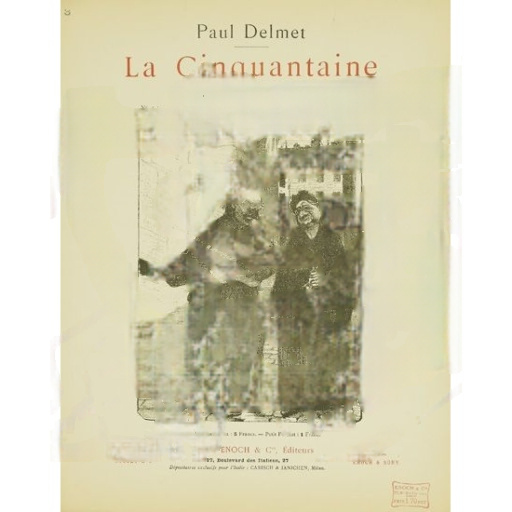}
        & \qualimage{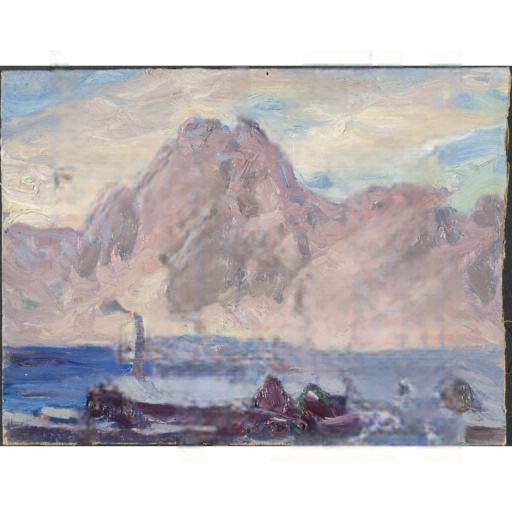}
        & \qualimage{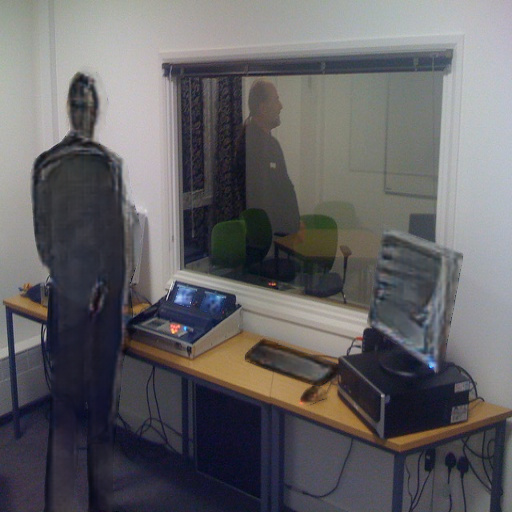}
        & \qualimage{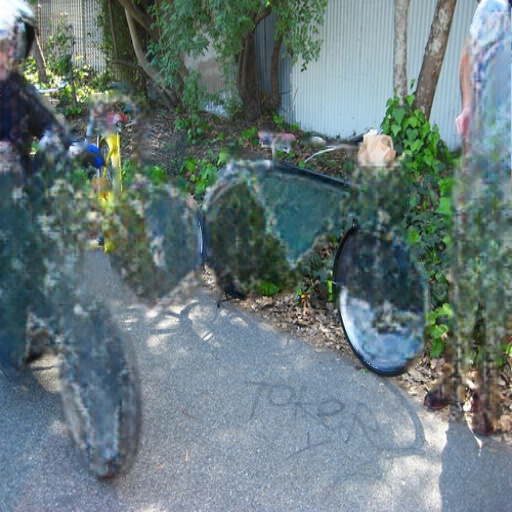}
        & \qualimage{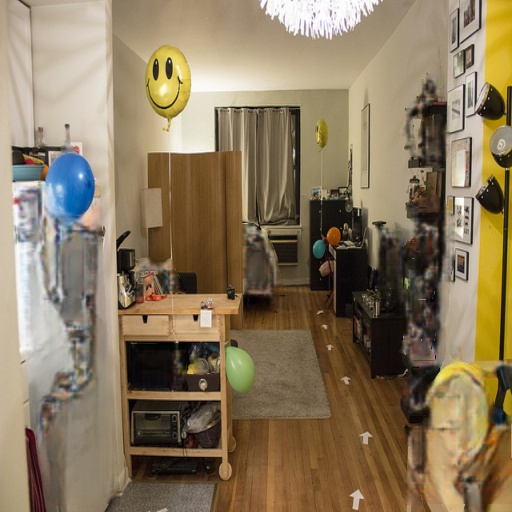} \\
        \qualmethodlabel{\quad OminiControl}
        & \qualimage{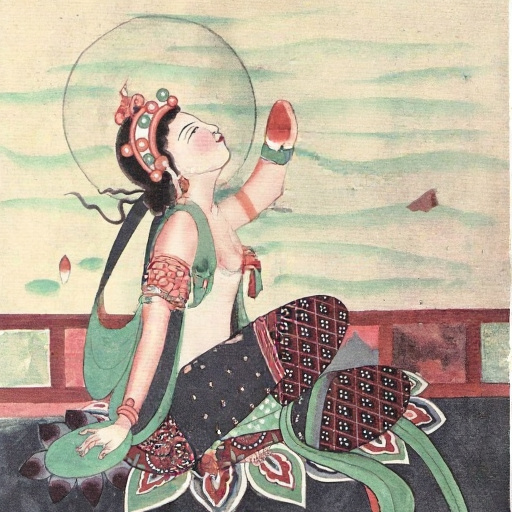}
        & \qualimage{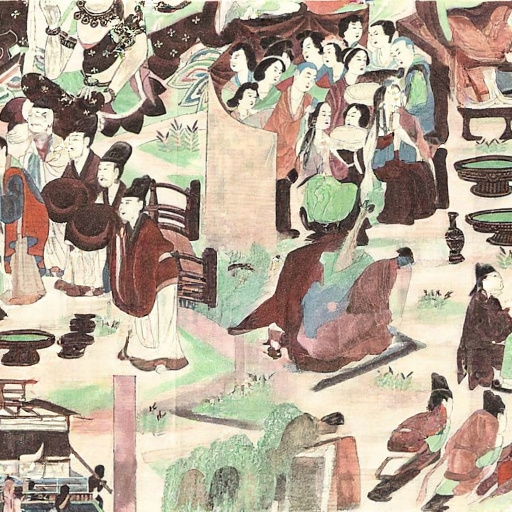}
        & \qualimage{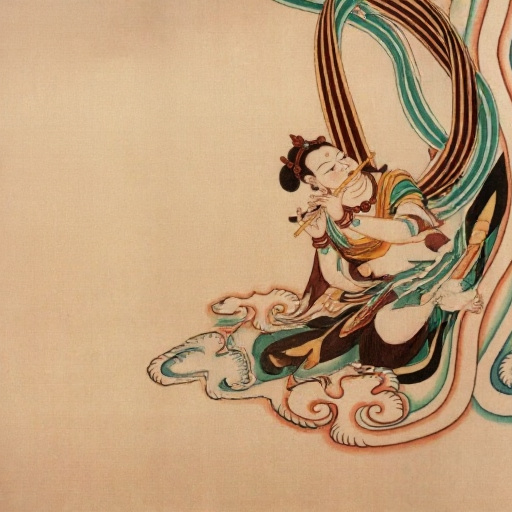}
        & \qualimage{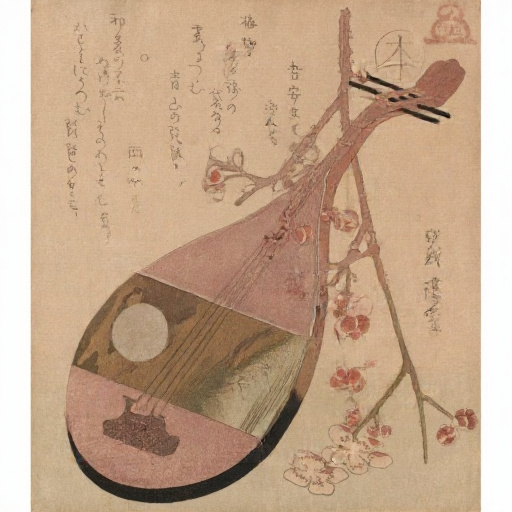}
        & \qualimage{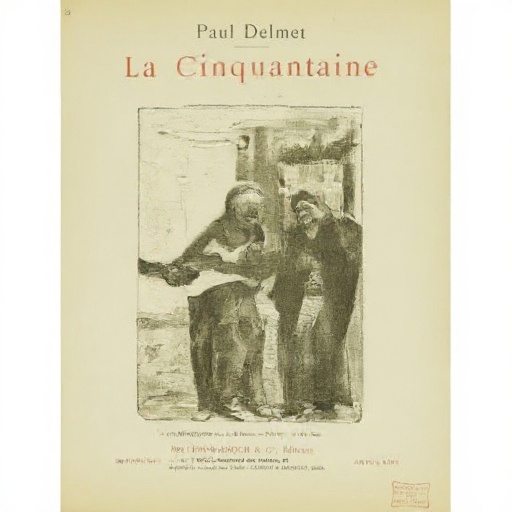}
        & \qualimage{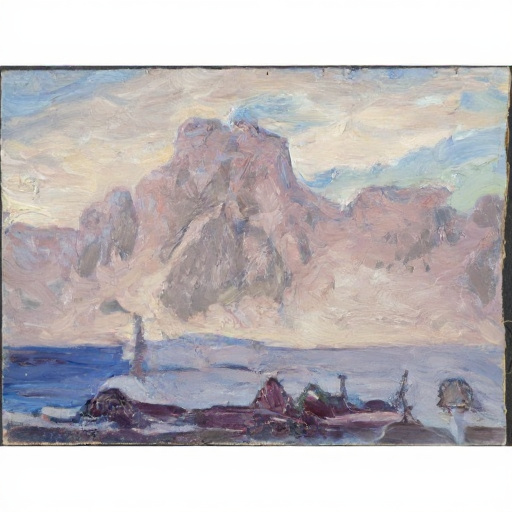}
        & \qualimage{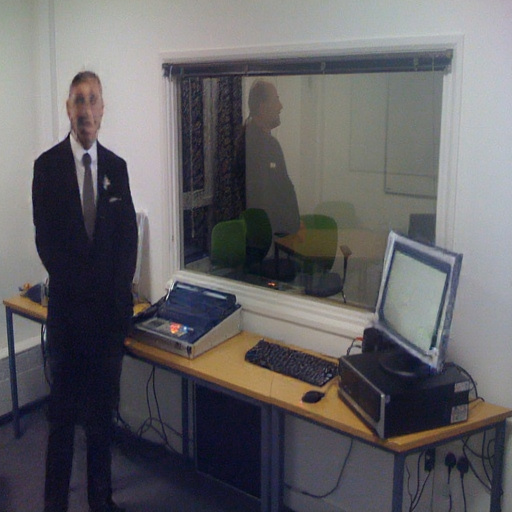}
        & \qualimage{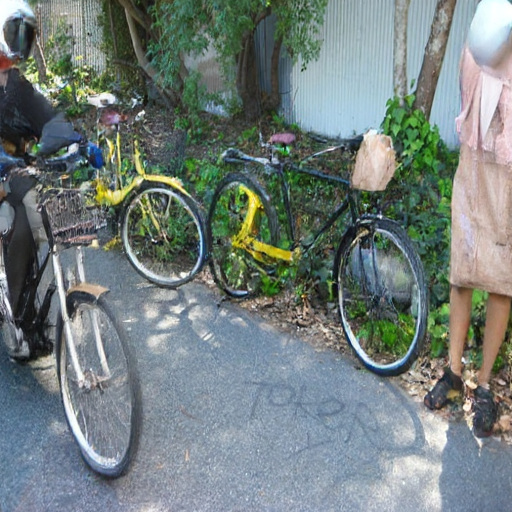}
        & \qualimage{} \\
        \hdashline
        \qualrowlabel{\quad Ours: RGB}
        & \qualimage{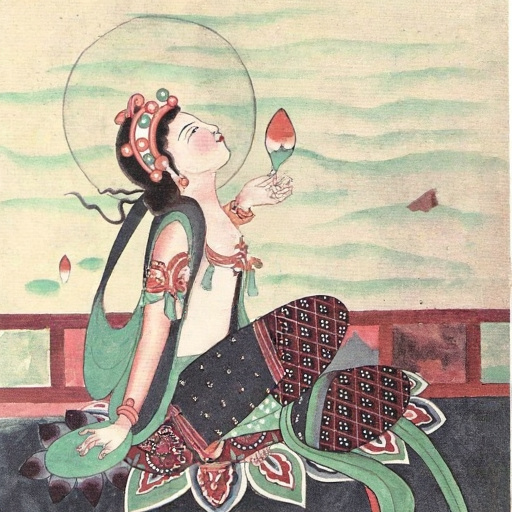}
        & \qualimage{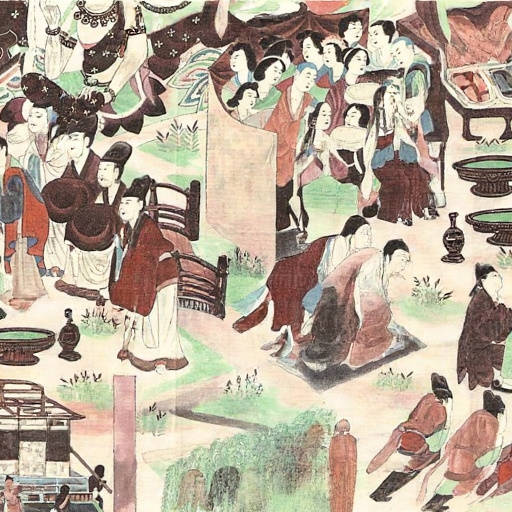}
        & \qualimage{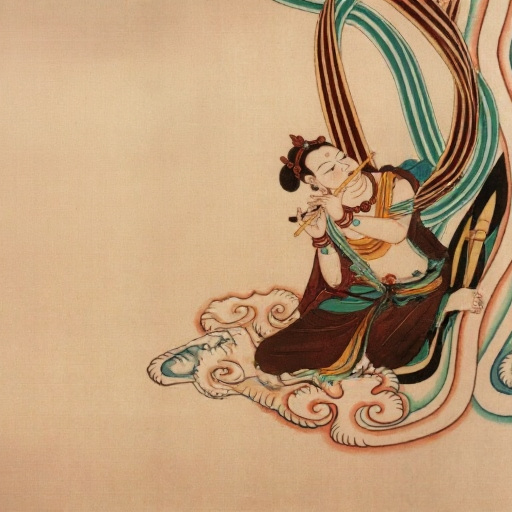}
        & \qualimage{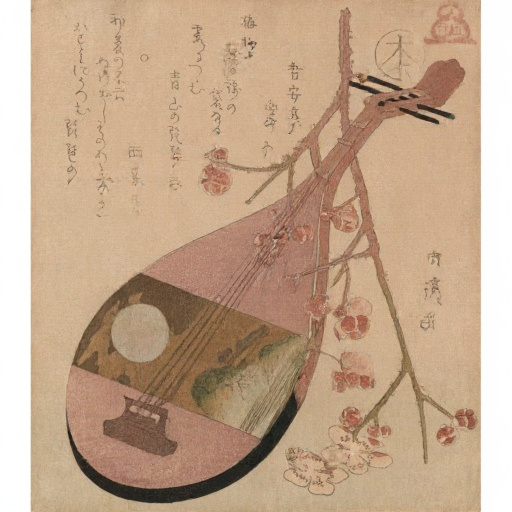}
        & \qualimage{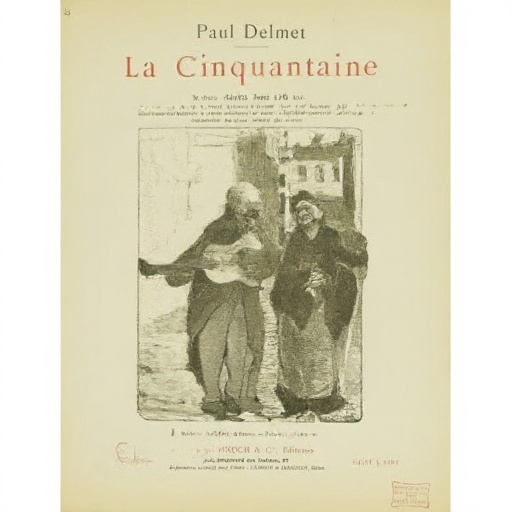}
        & \qualimage{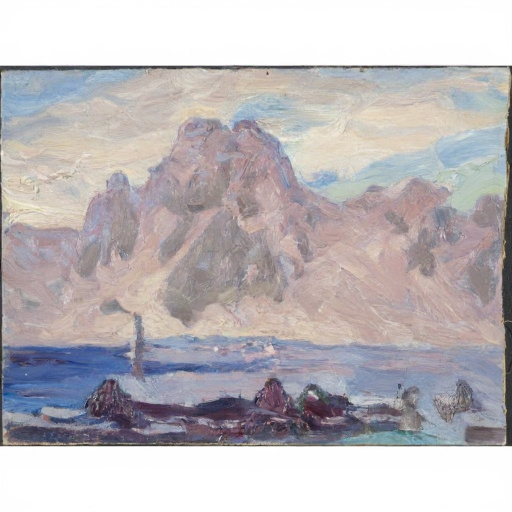}
        & \qualimage{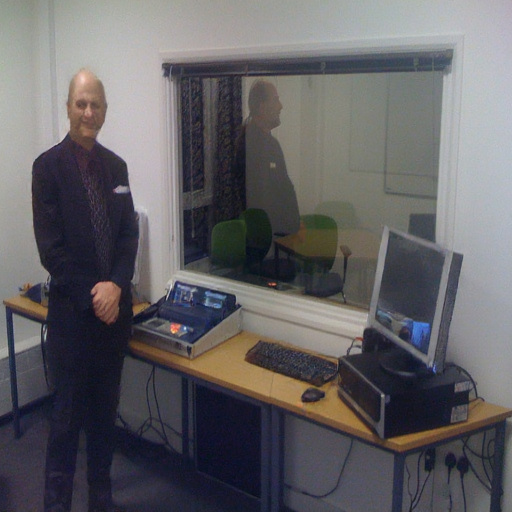}
        & \qualimage{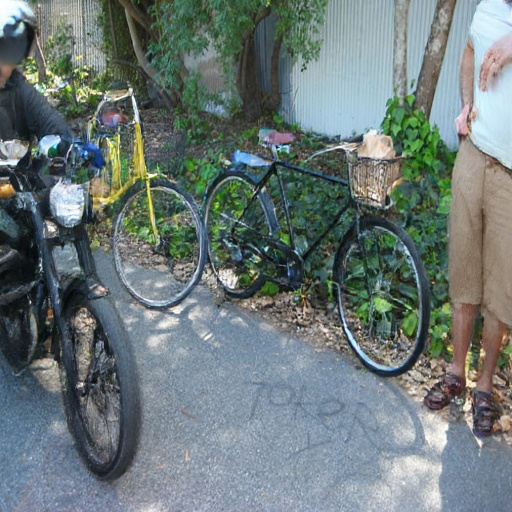}
        & \qualimage{} \\
        \qualrowlabel{\quad Ours: Refined structure}
        & \qualimage{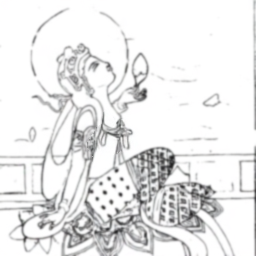}
        & \qualimage{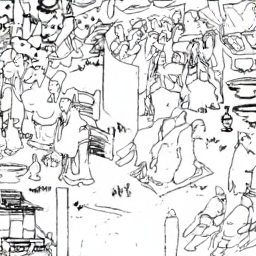}
        & \qualimage{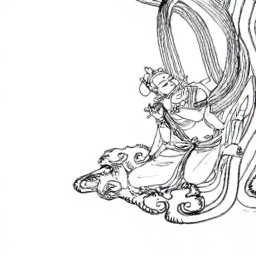}
        & \qualimage{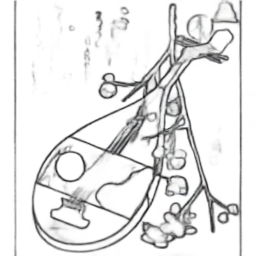}
        & \qualimage{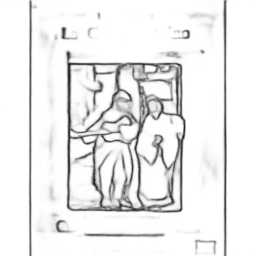}
        & \qualimage{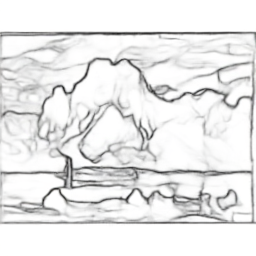}
        & \qualimage{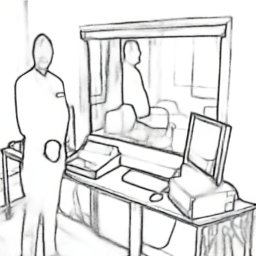}
        & \qualimage{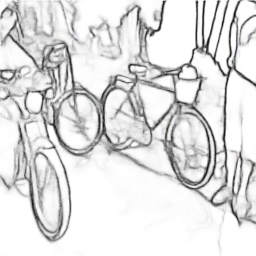}
        & \qualimage{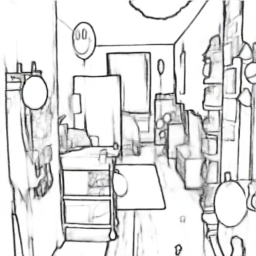} \\
        \hdashline
        \qualrowlabel{\quad GT: RGB}
        & \qualimage{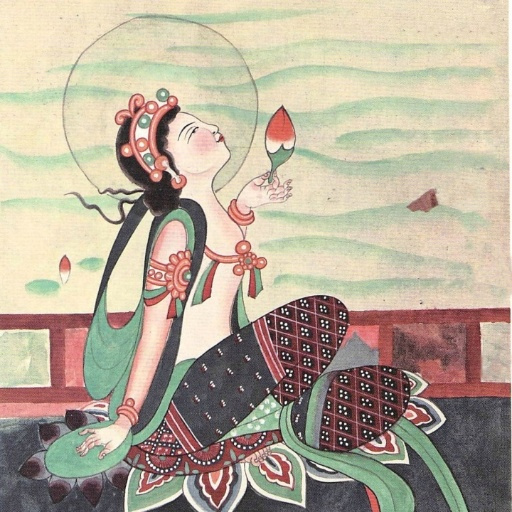}
        & \qualimage{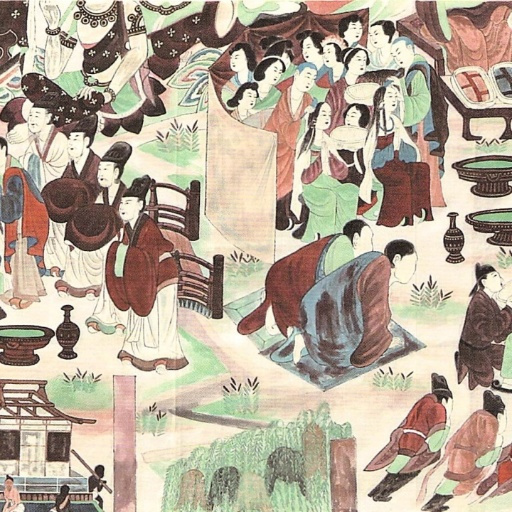}
        & \qualimage{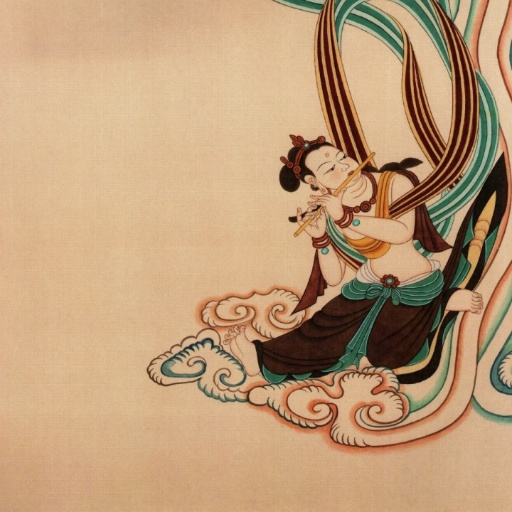}
        & \qualimage{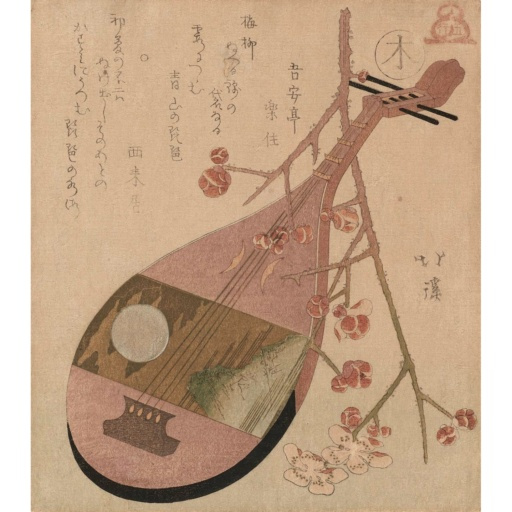}
        & \qualimage{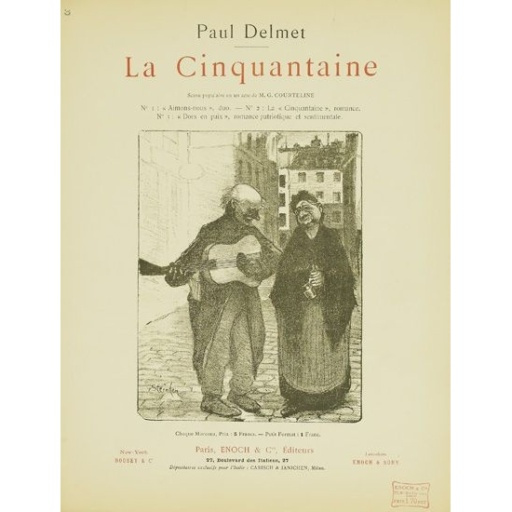}
        & \qualimage{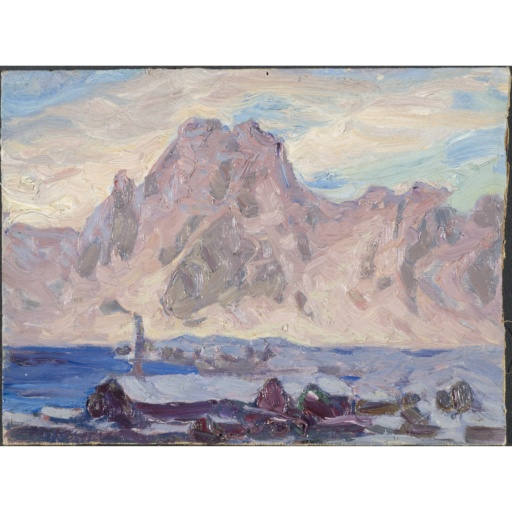}
        & \qualimage{}
        & \qualimage{}
        & \qualimage{} \\
        \qualrowlabel{\quad GT: Edge}
        & \qualimage{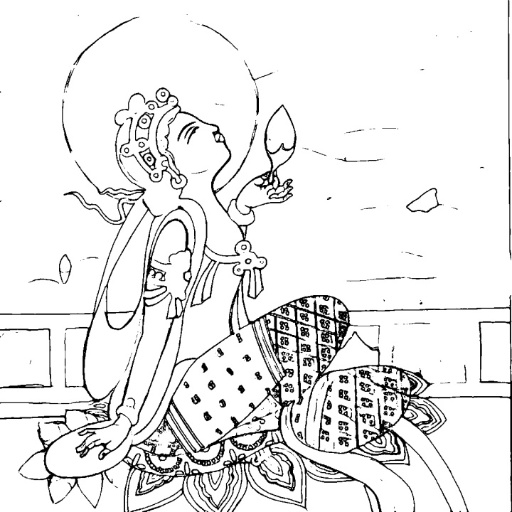}
        & \qualimage{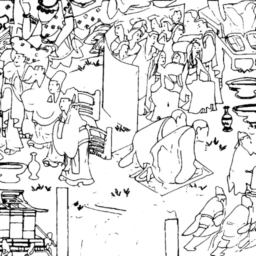}
        & \qualimage{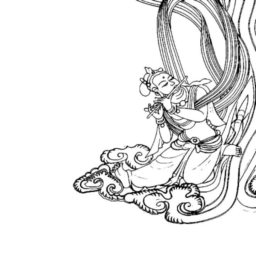}
        & \qualimage{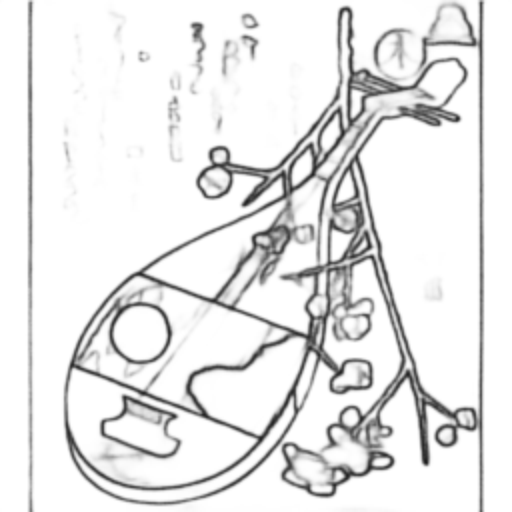}
        & \qualimage{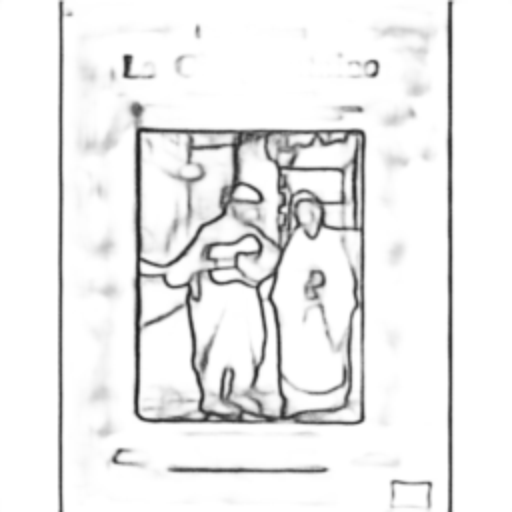}
        & \qualimage{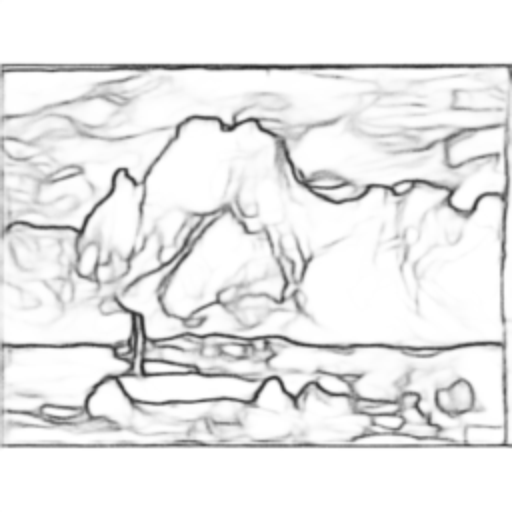}
        & \qualimage{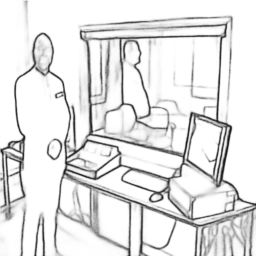}
        & \qualimage{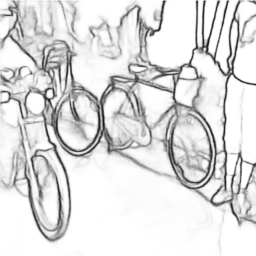}
        & \qualimage{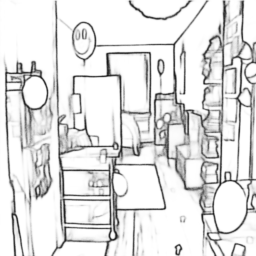} \\
        \bottomrule
    \end{tabular}
    \caption{Qualitative comparison on matched examples from DhMural1714,
    ArtBench, and COCO. Sketches are faded outside the inpainting mask for
    visualization; SketchSense additionally outputs refined structure.}
    \label{fig:qualitative_comparison}
\end{figure*}

\subsection{Experimental Setup}
\label{sec:experimental_setup}

We train on ArtBench~\cite{Liao2022ArtBench}, COCO~\cite{Lin2014COCO}, and
\mbox{DhMural1714}~\cite{Li2024Mural}, sampled with probabilities $0.3$, $0.6$, and
$0.1$, respectively. SketchSense builds on FLUX.1 Fill
[dev]~\cite{BlackForestLabs2024FluxFill} and operates at $512\times512$.
Training takes approximately seven days on one NVIDIA A100 using BF16, a batch
size of 12, and Adafactor with a learning rate of $10^{-5}$; inference uses a
guidance scale of 25. All methods use matched input conditions, and PSNR,
SSIM~\cite{Wang2004SSIM}, and LPIPS~\cite{Zhang2018LPIPS} are computed within
the inpainting mask.

\subsection{Quantitative Results}
\label{sec:quantitative_results}

\subsubsection{Comparison with Existing Methods}
Table~\ref{tab:main_quantitative_difficulty} shows that SketchSense consistently
outperforms the compared methods across the Easy, Medium, and Hard subsets.
The gains persist as difficulty increases, while the variants with and without
the signed prior achieve similar aggregate RGB metrics.

\begin{table*}[t!]
    \centering
    \scriptsize
    \setlength{\tabcolsep}{3.6pt}
    \begin{tabular}{lccccccccc}
        \toprule
        & \multicolumn{3}{c}{Easy}
        & \multicolumn{3}{c}{Medium}
        & \multicolumn{3}{c}{Hard} \\
        \cmidrule(lr){2-4}\cmidrule(lr){5-7}\cmidrule(lr){8-10}
        Method
        & PSNR $\uparrow$ & SSIM $\uparrow$ & LPIPS $\downarrow$
        & PSNR $\uparrow$ & SSIM $\uparrow$ & LPIPS $\downarrow$
        & PSNR $\uparrow$ & SSIM $\uparrow$ & LPIPS $\downarrow$ \\
        \midrule
        PDDP~\cite{Sharma2024SketchGuided}
        & 12.714 & 0.7820 & 0.2050
        & 11.350 & 0.6533 & 0.3019
        & 10.450 & 0.5237 & 0.3883 \\
        MaGIC (adapted)~\cite{Wang2024MaGIC}
        & 15.972 & 0.8176 & 0.1306
        & 14.402 & 0.6994 & 0.2029
        & 12.712 & 0.5658 & 0.2876 \\
        SketchRefiner (adapted)~\cite{Liu2024SketchRefiner}
        & 17.602 & 0.8359 & 0.1285
        & 16.156 & 0.7285 & 0.2034
        & 14.386 & 0.5916 & 0.3020 \\
        OminiControl~\cite{Tan2025OminiControl}
        & 18.103 & 0.8431 & 0.0921
        & 16.083 & 0.7274 & 0.1609
        & 13.902 & 0.5824 & 0.2479 \\
        \midrule
        SketchSense without signed prior
        & 19.428 & 0.8584 & 0.0720
        & 17.301 & 0.7502 & 0.1282
        & 14.933 & 0.6121 & 0.2040 \\
        SketchSense with signed prior
        & 19.442 & 0.8587 & 0.0723
        & 17.290 & 0.7506 & 0.1299
        & 14.898 & 0.6128 & 0.2061 \\
        \bottomrule
    \end{tabular}
    \caption{Quantitative comparison across the Easy, Medium, and Hard subsets.
    All metrics are computed within the inpainting mask.}
    \label{tab:main_quantitative_difficulty}
\end{table*}

\subsection{Qualitative Results}
\label{sec:qualitative_results}

\subsubsection{Comparison with Existing Methods}

Figure~\ref{fig:qualitative_comparison} shows that SketchSense produces more
coherent RGB restorations and refined structures than the compared methods
across DhMural1714, ArtBench, and COCO.
The improvement is especially visible in large or structurally complex holes,
where the recovered appearance remains consistent with the predicted structure.

\FloatBarrier

\subsubsection{Effect of Local Region Guidance}

\begin{figure}[t!]
    \centering
    \scriptsize
    \newcommand{\regionimage}[1]{%
        \includegraphics[width=.205\columnwidth,height=.205\columnwidth,keepaspectratio]{#1}}
    \newcommand{\regionboxedimage}[1]{%
        \begin{overpic}[width=.205\columnwidth,height=.205\columnwidth,keepaspectratio]{#1}
            \put(24,66){\color{blue!75!black}\linethickness{0.7pt}\framebox(58,20){}}
        \end{overpic}}
    \setlength{\tabcolsep}{0.5pt}
    \begin{tabular}{@{}cccc@{}}
        Masked input & Sketch & Signed region & Ground truth \\
        \regionimage{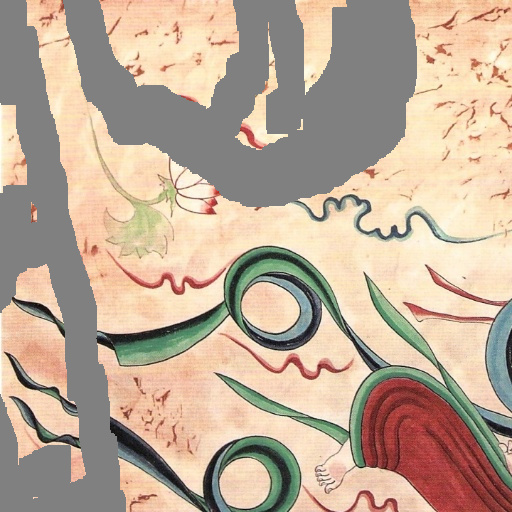} &
        \regionimage{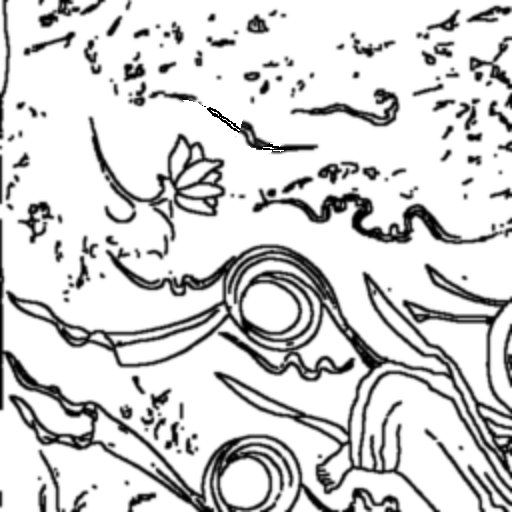} &
        \regionimage{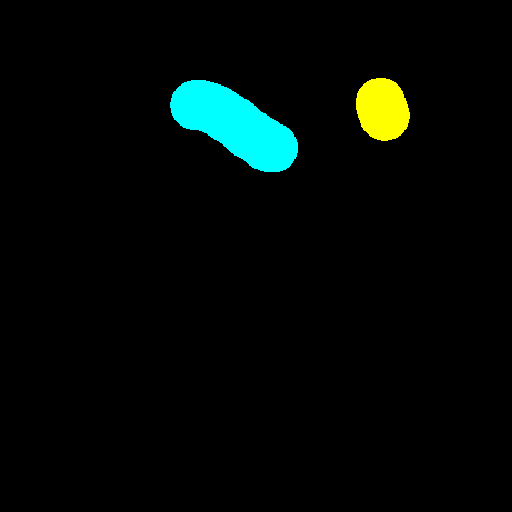} &
        \regionimage{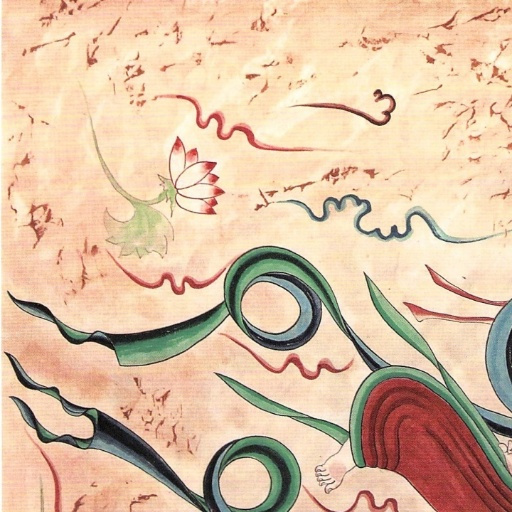}
    \end{tabular}

    \vspace{2pt}
    \setlength{\tabcolsep}{0.8pt}
    \begin{tabular}{@{}lcc@{}}
        & Without region & With region \\
        Restored RGB &
        \regionboxedimage{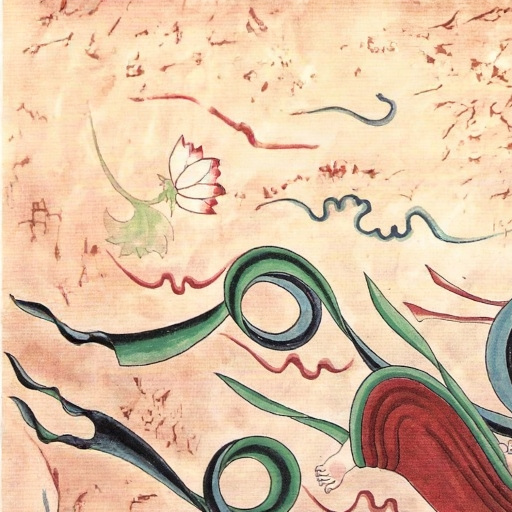} &
        \regionboxedimage{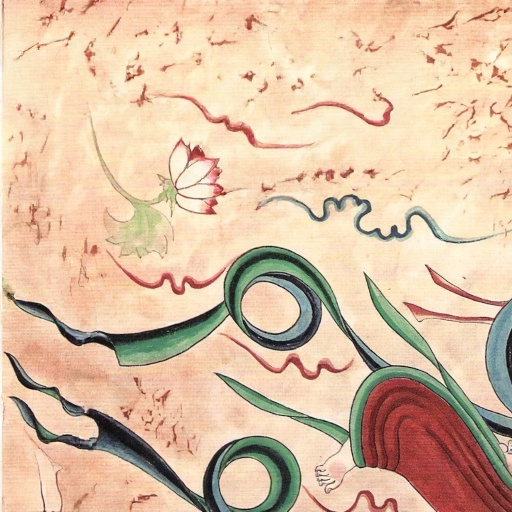} \\
        Refined structure &
        \regionboxedimage{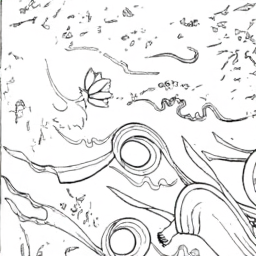} &
        \regionboxedimage{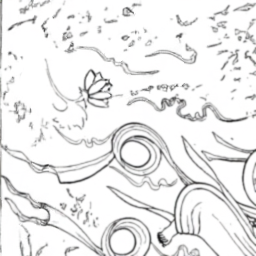}
    \end{tabular}
    \caption{Effect of local signed region guidance. The upper row shows the
    masked input, sketch, signed region map, and ground truth; yellow indicates
    sketch preservation and cyan permits contextual correction. The lower grid
    compares restored RGB and refined structure, with blue boxes marking the
    region of interest.}
    \label{fig:region_guidance_effect}
\end{figure}

Figure~\ref{fig:region_guidance_effect} illustrates how local signed guidance
changes the interpretation of selected sketch content while preserving the
overall composition. The corresponding RGB and refined-structure outputs show
that the resulting changes remain localized near the indicated structures.

\subsubsection{Same Image with Different Sketch Inputs}

\begin{figure}[t!]
    \centering
    \scriptsize
    \newcommand{\sketchboxedimage}[1]{%
        \begin{overpic}[width=.225\columnwidth]{#1}
            \put(40,20){\color{blue!75!black}\linethickness{0.7pt}\framebox(33,33){}}
        \end{overpic}}
    \setlength{\tabcolsep}{0.7pt}
    \begin{tabular}{@{}cccc@{}}
        \multicolumn{4}{c}{Sketch inputs} \\
        \sketchboxedimage{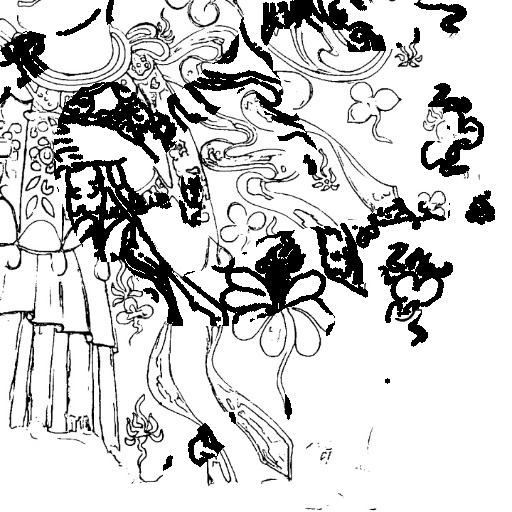} &
        \sketchboxedimage{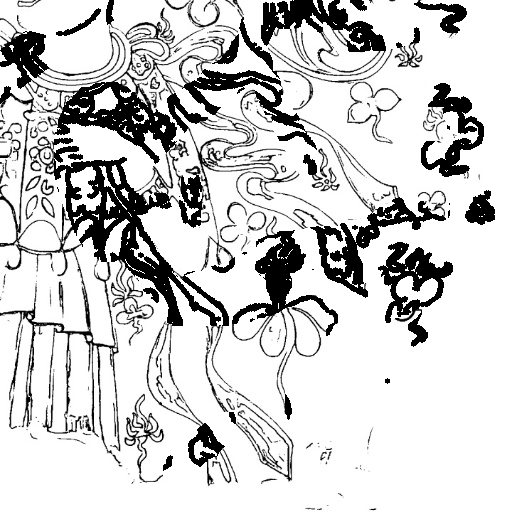} &
        \sketchboxedimage{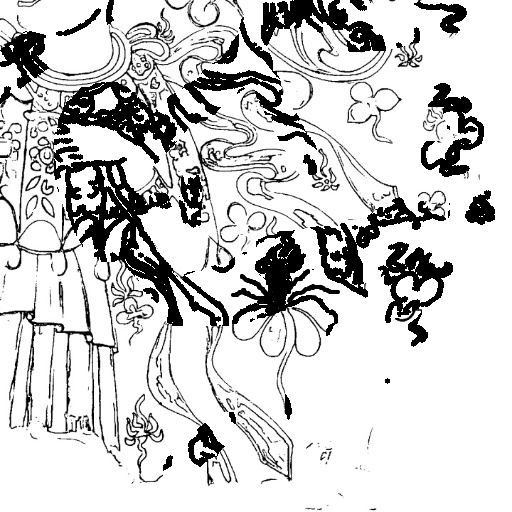} &
        \sketchboxedimage{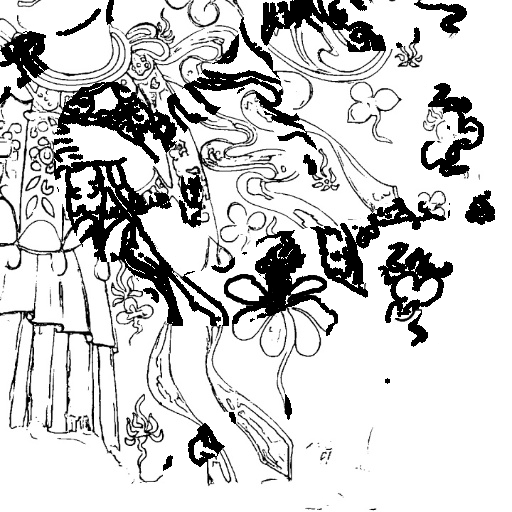} \\
        \multicolumn{4}{c}{Restored RGB} \\
        \sketchboxedimage{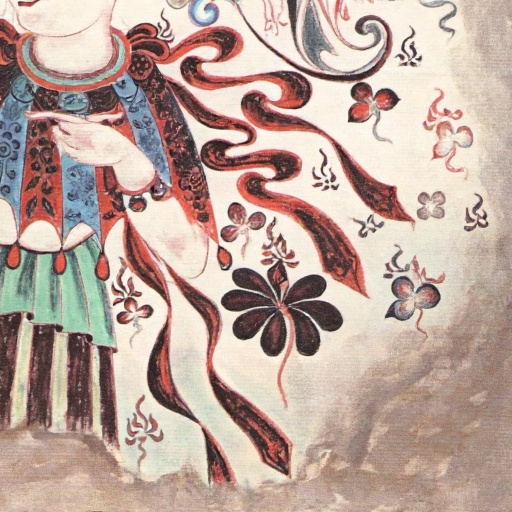} &
        \sketchboxedimage{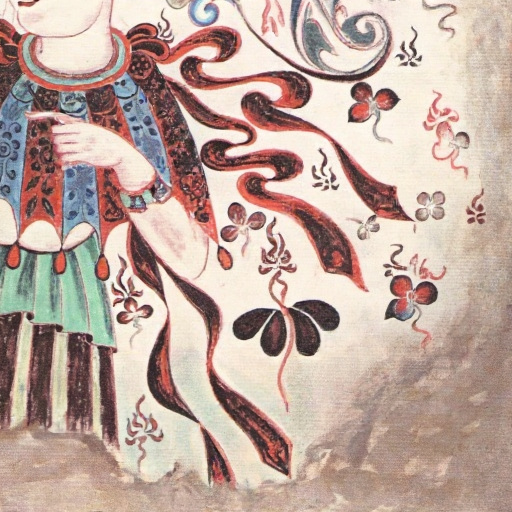} &
        \sketchboxedimage{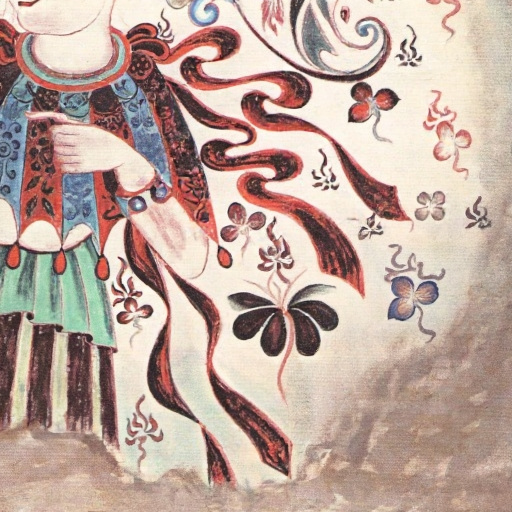} &
        \sketchboxedimage{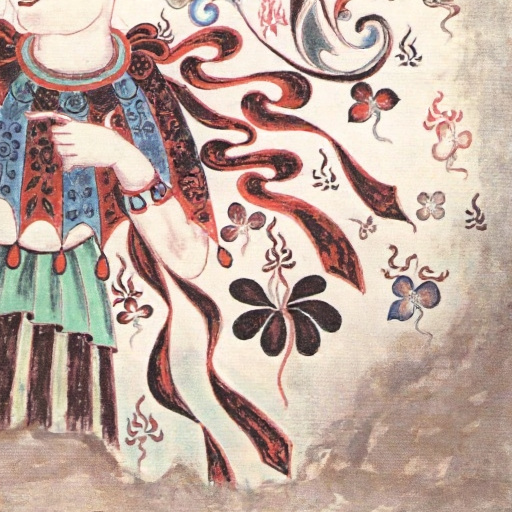}
    \end{tabular}
    \caption{Effect of different sketch inputs under a shared restoration
    setting. Each column pairs a sketch with its RGB restoration; blue boxes
    mark the shared region of interest.}
    \label{fig:different_sketch_inputs}
\end{figure}

Figure~\ref{fig:different_sketch_inputs} varies the sketch within a shared
restoration setting. Local stroke differences produce distinct yet coherent RGB
completions, demonstrating that SketchSense responds to the supplied structural
evidence.

\subsection{Ablation Studies}
\label{sec:ablation_studies}

\subsubsection{Component Ablation}

Table~\ref{tab:component_ablation} compares the complete model with component
removals and a sequential variant that fixes the recovered structure before
restoring RGB. SketchSense achieves
the best SSIM, LPIPS, and Edge F1. Removing intrinsic regulation slightly raises
PSNR but degrades the perceptual and structural metrics.

\begin{table}[t!]
    \centering
    \scriptsize
    \setlength{\tabcolsep}{1.5pt}
    \renewcommand{\arraystretch}{1.04}
    \begin{tabular}{@{}p{.42\columnwidth}cccc@{}}
        \toprule
        Variant & PSNR $\uparrow$ & SSIM $\uparrow$ & LPIPS $\downarrow$
        & Edge F1@2px $\uparrow$ \\
        \midrule
        w/o cross-branch fusion             & 15.61 & 0.6531 & 0.2260 & 0.6882 \\
        Sequential RGB--structure recovery  & 17.27 & 0.6828 & 0.1746 & 0.6882 \\
        w/o intrinsic spatial regulation    & \textbf{17.50} & 0.6736 & 0.2242 & 0.7325 \\
        w/o explicit signed modeling        & 17.23 & 0.6835 & 0.1720 & 0.7398 \\
        \midrule
        SketchSense                         & 17.49 & \textbf{0.6848} & \textbf{0.1707} & \textbf{0.7469} \\
        \bottomrule
    \end{tabular}
    \caption{Component ablation. All metrics are computed inside the inpainting
    hole.}
    \label{tab:component_ablation}
\end{table}


\subsubsection{Signed-Region Breakdown}

Table~\ref{tab:signed_region_ablation} compares SketchSense with a variant
without explicit signed modeling on balanced neutral, positive, and negative
conditions, corresponding to no prior, faithful sketch use, and contextual
correction. The explicit prior leaves neutral results nearly unchanged while
improving Region F1 under both positive and negative guidance.

\begin{table}[t!]
    \centering
    \scriptsize
    \setlength{\tabcolsep}{1.1pt}
    \renewcommand{\arraystretch}{1.04}
    \begin{tabular}{@{}p{.14\columnwidth}p{.25\columnwidth}cccc@{}}
        \toprule
        Mode & Variant & PSNR $\uparrow$ & SSIM $\uparrow$ & LPIPS $\downarrow$
        & Region F1@2px $\uparrow$ \\
        \midrule
        Neutral  & w/o explicit & \textbf{17.02} & \textbf{0.6851} & \textbf{0.1733} & 0.7019 \\
                 & SketchSense  & \textbf{17.02} & 0.6850 & \textbf{0.1733} & \textbf{0.7047} \\
        Positive & w/o explicit & 17.69 & 0.6949 & 0.1627 & 0.7344 \\
                 & SketchSense  & \textbf{17.77} & \textbf{0.6971} & \textbf{0.1611} & \textbf{0.8107} \\
        Negative & w/o explicit & 16.97 & 0.6702 & 0.1805 & 0.5935 \\
                 & SketchSense  & \textbf{17.03} & \textbf{0.6715} & \textbf{0.1788} & \textbf{0.6266} \\
        \bottomrule
    \end{tabular}
    \caption{Results by signed-region mode. RGB metrics are measured in the
    hole, and Region F1@2px in the evaluated region.}
    \label{tab:signed_region_ablation}
\end{table}

\FloatBarrier

\subsubsection{Spatial Gate Analysis}

\begin{figure}[!htbp]
    \centering
    \includegraphics[page=3,width=\columnwidth]{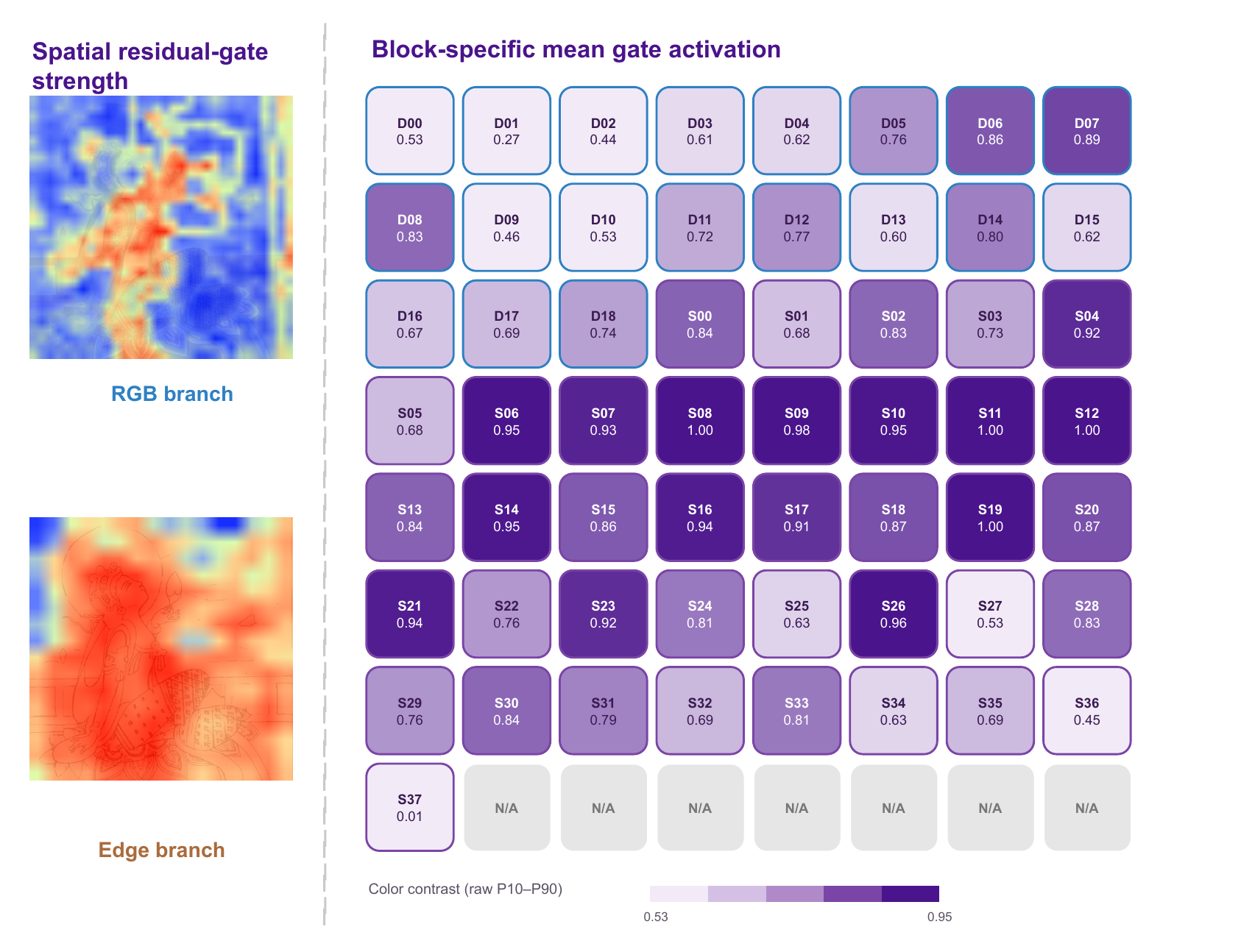}
    \caption{Representative spatial residual-gate visualization at denoising
    step 35. Gate activation varies across spatial locations, branches, and
    transformer blocks.}
    \label{fig:spatial_gate_diagnostic}
\end{figure}

To inspect whether cross-branch residual injection varies as designed, we
visualize one representative held-out example at denoising step 35 in
Fig.~\ref{fig:spatial_gate_diagnostic}. The maps and blockwise means exhibit
both spatial and block-dependent variation, consistent with the intended
block-specific gating. Because this is a single-example diagnostic, we do not
treat it as aggregate performance evidence.

\section{Conclusion}
\label{sec:conclusion}

We presented SketchSense, a sketch-guided inpainting framework that jointly
restores RGB content and interprets imperfect sketch guidance. Synchronous RGB
and structure streams exchange evidence through Bidirectional Attention Fusion,
while sketch-aware spatial regulation adapts sketch use to local generation
states. An optional signed prior conveys local preserve-or-correct intent, and
the refined structure makes the model's interpretation observable. Experiments
on natural images and complex artwork show consistent improvements over
existing inpainting and sketch-conditioned methods across all evaluated
difficulty groups. Future work will focus on more efficient dual-stream
modeling, higher-resolution restoration, and simpler interaction for specifying
uncertain sketch regions.

{
    \small
    \bibliographystyle{ieeenat_fullname}
    \bibliography{main}

@article{Liu2024SketchRefiner,
  title   = {Toward Interactive Image Inpainting via Robust Sketch Refinement},
  author  = {Liu, Chang and Xu, Shunxin and Peng, Jialun and Zhang, Kaidong and Liu, Dong},
  journal = {IEEE Transactions on Multimedia},
  volume  = {26},
  pages   = {9973--9987},
  year    = {2024},
  doi     = {10.1109/TMM.2024.3402620},
  url     = {https://doi.org/10.1109/TMM.2024.3402620}
}

@inproceedings{Sharma2024SketchGuided,
  title     = {Sketch-guided Image Inpainting with Partial Discrete Diffusion Process},
  author    = {Sharma, Nakul and Tripathi, Aditay and Chakraborty, Anirban and Mishra, Anand},
  booktitle = {Proceedings of the IEEE/CVF Conference on Computer Vision and Pattern Recognition (CVPR) Workshops},
  month     = {June},
  year      = {2024},
  pages     = {6024--6034}
}

@inproceedings{Wang2024MaGIC,
  title     = {{MaGIC}: Multi-modality Guided Image Completion},
  author    = {Wang, Hao and Yu, Yongsheng and Luo, Tiejian and Fan, Heng and Zhang, Libo},
  booktitle = {The Twelfth International Conference on Learning Representations},
  year      = {2024},
  url       = {https://openreview.net/forum?id=o7x0XVlCpX}
}

@inproceedings{Xie2023SmartBrush,
  title     = {SmartBrush: Text and Shape Guided Object Inpainting With Diffusion Model},
  author    = {Xie, Shaoan and Zhang, Zhifei and Lin, Zhe and Hinz, Tobias and Zhang, Kun},
  booktitle = {Proceedings of the IEEE/CVF Conference on Computer Vision and Pattern Recognition (CVPR)},
  month     = {June},
  year      = {2023},
  pages     = {22428--22437},
  doi       = {10.1109/CVPR52729.2023.02148}
}

@inproceedings{Liu2024StructureMatters,
  title     = {Structure Matters: Tackling the Semantic Discrepancy in Diffusion Models for Image Inpainting},
  author    = {Liu, Haipeng and Wang, Yang and Qian, Biao and Wang, Meng and Rui, Yong},
  booktitle = {Proceedings of the IEEE/CVF Conference on Computer Vision and Pattern Recognition (CVPR)},
  month     = {June},
  year      = {2024},
  pages     = {8038--8047},
  doi       = {10.1109/CVPR52733.2024.00768}
}

@article{Li2024Mural,
  title   = {Line Drawing-Guided Progressive Inpainting for Mural Damage},
  author  = {Li, Luxi and Zou, Qin and Zhang, Fan and Yu, Hongkai and Chen, Long and Song, Chengfang and Huang, Xianfeng and Wang, Xiaoguang and Li, Qingquan},
  journal = {Journal on Computing and Cultural Heritage},
  volume  = {18},
  number  = {3},
  pages   = {1--20},
  month   = {September},
  year    = {2025},
  doi     = {10.1145/3737703},
  url     = {https://doi.org/10.1145/3737703},
  publisher = {Association for Computing Machinery}
}

@misc{Cao2023ZITSpp,
  title         = {ZITS++: Image Inpainting by Improving the Incremental Transformer on Structural Priors},
  author        = {Cao, Chenjie and Dong, Qiaole and Fu, Yanwei},
  year          = {2023},
  eprint        = {2210.05950},
  archivePrefix = {arXiv},
  primaryClass  = {cs.CV},
  url           = {https://arxiv.org/abs/2210.05950}
}

@inproceedings{Cohen2025ConditionalBalance,
  title     = {Conditional Balance: Improving Multi-Conditioning Trade-Offs in Image Generation},
  author    = {Cohen, Nadav Z. and Nir, Oron and Shamir, Ariel},
  booktitle = {Proceedings of the IEEE/CVF Conference on Computer Vision and Pattern Recognition (CVPR)},
  month     = {June},
  year      = {2025},
  pages     = {2641--2650}
}

@inproceedings{Huang2025MTADiffusion,
  title     = {MTADiffusion: Mask Text Alignment Diffusion Model for Object Inpainting},
  author    = {Huang, Jun and Liu, Ting and Wu, Yihang and Qu, Xiaochao and Liu, Luoqi and Hu, Xiaolin},
  booktitle = {Proceedings of the IEEE/CVF Conference on Computer Vision and Pattern Recognition (CVPR)},
  month     = {June},
  pages     = {18325--18334},
  year      = {2025}
}

@inproceedings{Shen2024SCFG,
  title     = {Rethinking the Spatial Inconsistency in Classifier-Free Diffusion Guidance},
  author    = {Shen, Dazhong and Song, Guanglu and Xue, Zeyue and Wang, Fu-Yun and Liu, Yu},
  booktitle = {Proceedings of the IEEE/CVF Conference on Computer Vision and Pattern Recognition (CVPR)},
  month     = {June},
  pages     = {9370--9379},
  year      = {2024}
}

@inproceedings{Tan2025OminiControl,
  title     = {OminiControl: Minimal and Universal Control for Diffusion Transformer},
  author    = {Tan, Zhenxiong and Liu, Songhua and Yang, Xingyi and Xue, Qiaochu and Wang, Xinchao},
  booktitle = {Proceedings of the IEEE/CVF International Conference on Computer Vision (ICCV)},
  month     = {October},
  pages     = {14940--14950},
  year      = {2025}
}

@inproceedings{Phung2024AttentionRefocusing,
  title     = {Grounded Text-to-Image Synthesis with Attention Refocusing},
  author    = {Phung, Quynh and Ge, Songwei and Huang, Jia-Bin},
  booktitle = {Proceedings of the IEEE/CVF Conference on Computer Vision and Pattern Recognition (CVPR)},
  month     = {June},
  pages     = {7932--7942},
  year      = {2024}
}

@inproceedings{Zhang2023ControlNet,
  title     = {Adding Conditional Control to Text-to-Image Diffusion Models},
  author    = {Zhang, Lvmin and Rao, Anyi and Agrawala, Maneesh},
  booktitle = {Proceedings of the IEEE/CVF International Conference on Computer Vision (ICCV)},
  month     = {October},
  pages     = {3836--3847},
  year      = {2023}
}

@inproceedings{Kim2023DenseDiffusion,
  title     = {Dense Text-to-Image Generation with Attention Modulation},
  author    = {Kim, Yunji and Lee, Jiyoung and Kim, Jin-Hwa and Ha, Jung-Woo and Zhu, Jun-Yan},
  booktitle = {Proceedings of the IEEE/CVF International Conference on Computer Vision (ICCV)},
  month     = {October},
  pages     = {7701--7711},
  year      = {2023}
}

@inproceedings{Zhang2023JointNet,
  title     = {{JointNet}: Extending Text-to-Image Diffusion for Dense Distribution Modeling},
  author    = {Zhang, Jingyang and Li, Shiwei and Lu, Yuanxun and Fang, Tian and McKinnon, David Neil and Tsin, Yanghai and Quan, Long and Yao, Yao},
  booktitle = {The Twelfth International Conference on Learning Representations},
  year      = {2024},
  url       = {https://openreview.net/forum?id=kv5xE1p3jz}
}

@inproceedings{Kwon2025JointDiT,
  title     = {{JointDiT}: Enhancing {RGB}-Depth Joint Modeling with Diffusion Transformers},
  author    = {Kwon, Byung-Ki and Dai, Qi and Lee, Hyoseok and Luo, Chong and Oh, Tae-Hyun},
  booktitle = {Proceedings of the IEEE/CVF International Conference on Computer Vision (ICCV)},
  month     = {October},
  year      = {2025},
  pages     = {25261--25271}
}

@misc{BlackForestLabs2024FluxFill,
  title        = {{FLUX.1 Fill [dev]}},
  author       = {{Black Forest Labs}},
  year         = {2024},
  howpublished = {Hugging Face model card},
  url          = {https://huggingface.co/black-forest-labs/FLUX.1-Fill-dev}
}

@inproceedings{Lipman2023FlowMatching,
  title     = {Flow Matching for Generative Modeling},
  author    = {Lipman, Yaron and Chen, Ricky T. Q. and Ben-Hamu, Heli and Nickel, Maximilian and Le, Matt},
  booktitle = {The Eleventh International Conference on Learning Representations},
  year      = {2023},
  url       = {https://openreview.net/forum?id=PqvMRDCJT9t}
}

@inproceedings{Esser2024RectifiedFlowTransformer,
  title     = {Scaling Rectified Flow Transformers for High-Resolution Image Synthesis},
  author    = {Esser, Patrick and Kulal, Sumith and Blattmann, Andreas and Entezari, Rahim and M{\"u}ller, Jonas and Saini, Harry and Levi, Yam and Lorenz, Dominik and Sauer, Axel and Boesel, Frederic and Podell, Dustin and Dockhorn, Tim and English, Zion and Rombach, Robin},
  booktitle = {Proceedings of the 41st International Conference on Machine Learning},
  pages     = {12606--12633},
  year      = {2024},
  editor    = {Salakhutdinov, Ruslan and Kolter, Zico and Heller, Katherine and Weller, Adrian and Oliver, Nuria and Scarlett, Jonathan and Berkenkamp, Felix},
  volume    = {235},
  series    = {Proceedings of Machine Learning Research},
  month     = {21--27 Jul},
  publisher = {PMLR},
  url       = {https://proceedings.mlr.press/v235/esser24a.html}
}

@inproceedings{Hu2022LoRA,
  title     = {Lo{RA}: Low-Rank Adaptation of Large Language Models},
  author    = {Hu, Edward J. and Shen, Yelong and Wallis, Phillip and Allen-Zhu, Zeyuan and Li, Yuanzhi and Wang, Shean and Wang, Lu and Chen, Weizhu},
  booktitle = {International Conference on Learning Representations},
  year      = {2022},
  url       = {https://openreview.net/forum?id=nZeVKeeFYf9}
}

@misc{Liao2022ArtBench,
  title         = {The {ArtBench} Dataset: Benchmarking Generative Models with Artworks},
  author        = {Liao, Peiyuan and Li, Xiuyu and Liu, Xihui and Keutzer, Kurt},
  year          = {2022},
  eprint        = {2206.11404},
  archivePrefix = {arXiv},
  primaryClass  = {cs.CV},
  url           = {https://arxiv.org/abs/2206.11404}
}

@inproceedings{Lin2014COCO,
  title     = {Microsoft {COCO}: Common Objects in Context},
  author    = {Lin, Tsung-Yi and Maire, Michael and Belongie, Serge and Hays, James and Perona, Pietro and Ramanan, Deva and Doll{\'a}r, Piotr and Zitnick, C. Lawrence},
  booktitle = {Computer Vision -- ECCV 2014},
  pages     = {740--755},
  year      = {2014},
  volume    = {8693},
  series    = {Lecture Notes in Computer Science},
  publisher = {Springer, Cham},
  doi       = {10.1007/978-3-319-10602-1_48},
  url       = {https://doi.org/10.1007/978-3-319-10602-1_48}
}

@article{Wang2004SSIM,
  title   = {Image Quality Assessment: From Error Visibility to Structural Similarity},
  author  = {Wang, Zhou and Bovik, Alan C. and Sheikh, Hamid R. and Simoncelli, Eero P.},
  journal = {IEEE Transactions on Image Processing},
  volume  = {13},
  number  = {4},
  pages   = {600--612},
  month   = {April},
  year    = {2004},
  doi     = {10.1109/TIP.2003.819861},
  url     = {https://doi.org/10.1109/TIP.2003.819861}
}

@inproceedings{Zhang2018LPIPS,
  title     = {The Unreasonable Effectiveness of Deep Features as a Perceptual Metric},
  author    = {Zhang, Richard and Isola, Phillip and Efros, Alexei A. and Shechtman, Eli and Wang, Oliver},
  booktitle = {Proceedings of the IEEE Conference on Computer Vision and Pattern Recognition (CVPR)},
  month     = {June},
  pages     = {586--595},
  year      = {2018}
}

@inproceedings{Zhou2024MuGE,
  title     = {{MuGE}: Multiple Granularity Edge Detection},
  author    = {Zhou, Caixia and Huang, Yaping and Pu, Mengyang and Guan, Qingji and Deng, Ruoxi and Ling, Haibin},
  booktitle = {Proceedings of the IEEE/CVF Conference on Computer Vision and Pattern Recognition (CVPR)},
  month     = {June},
  pages     = {25952--25962},
  year      = {2024},
  doi       = {10.1109/CVPR52733.2024.02452}
}
}
\end{document}